\documentclass{article} 
\usepackage{./template/iclr2021/iclr2021_conference,times}

\usepackage[utf8]{inputenc} 
\usepackage[T1]{fontenc}    
\usepackage{hyperref}       
\usepackage{url}            
\usepackage{booktabs}       
\usepackage{amsfonts}       
\usepackage{nicefrac}       
\usepackage{microtype}      

\usepackage{graphicx}
\graphicspath{{./figure/}}
\usepackage{bm}
\usepackage{amsmath}
\usepackage{multirow}
\usepackage{placeins}
\usepackage{subcaption}
\usepackage{comment}
\usepackage{xcolor,soul}
\usepackage{wrapfig}

\title{\mbox{Group-Connected Multilayer Perceptron Networks}}

\author{%
  Mohammad Kachuee, Sajad Darabi, Shayan Fazeli, Majid Sarrafzadeh \\
  Department of Computer Science \\
  University of California, Los Angeles (UCLA)\\ \\
}

\def\preprint{}

\ifdefined\preprint
    \iclrfinalcopy
\fi

\begin{document}

\maketitle

\ifdefined\preprint
  \lhead{}
\fi

\begin{abstract}
  Despite the success of deep learning in domains such as image, voice, and graphs, there has been little progress in deep representation learning for domains without a known structure between features. For instance, a tabular dataset of different demographic and clinical factors where the feature interactions are not given as a prior.
  In this paper, we propose Group-Connected Multilayer Perceptron (GMLP) networks to enable deep representation learning in these domains.
  GMLP is based on the idea of learning expressive feature combinations (groups) and exploiting them to reduce the network complexity by defining local group-wise operations. During the training phase, GMLP learns a sparse feature grouping matrix using temperature annealing softmax with an added entropy loss term to encourage the sparsity. Furthermore, an architecture is suggested which resembles binary trees, where group-wise operations are followed by pooling operations to combine information; reducing the number of groups as the network grows in depth.
  To evaluate the proposed method, we conducted experiments on different real-world datasets covering various application areas. Additionally, we provide visualizations on MNIST and synthesized data. According to the results, GMLP is able to successfully learn and exploit expressive feature combinations and achieve state-of-the-art classification performance on different datasets.
 
\end{abstract}

\section{Introduction}

Deep neural networks have been quite successful across various machine learning tasks. However, this advancement has been mostly limited to certain domains. For example in image and voice data, one can leverage domain properties such as location invariance, scale invariance, coherence, etc. via using convolutional layers~\citep{goodfellow2016deep}. Alternatively, for graph data, graph convolutional networks were suggested to leverage adjacency patterns present in datasets structured as a graph~\citep{kipf2016semi,xu2018how}.

However, there has been little progress in learning deep representations for datasets that do not follow a particular known structure in the feature domain. Take for instance the case of a simple tabular dataset for disease diagnosis. Such a dataset may consist of features from different categories such as demographics (e.g., age, gender, income, etc.), examinations (e.g., blood pressure, lab results, etc.), and other clinical conditions. In this scenario, the lack of any known structure between features to be used as a prior would lead to the use of a fully-connected multilayer perceptron network (MLP). Nonetheless, it has been known in the literature that MLP architectures, due to their huge complexity, do not usually admit efficient training and generalization for networks of more than a few layers.

In this paper, we propose Group-Connected Multilayer Perceptron (GMLP) networks. The main idea behind GMLP is to learn and leverage expressive feature subsets, henceforth referred to as \textit{feature groups}. A feature group is defined as a subset of features that provides a meaningful representation or high-level concept that would help the downstream task~\footnote{In this paper, the expression "group" is not related to the group in a mathematical sense, and it only represents a subset of features.}. For instance, in the disease diagnosis example, the combination of a certain blood factor and age might be the indicator of a higher level clinical condition which would help the final classification task. Furthermore, GMLP leverages feature groups limiting network connections to local group-wise connections and builds a feature hierarchy via merging groups as the network grows in depth. GMLP can be seen as an architecture that learns expressive feature combinations and leverages them via group-wise operations.

The main contributions of this paper are as follows: $(i)$ proposing a method for end-to-end learning of expressive feature combinations, $(ii)$ suggesting a network architecture to utilize feature groups and local connections to build deep representations, $(iii)$ conducting extensive experiments demonstrating the effectiveness of GMLP as well as visualizations and ablation studies for better understanding of the suggested architecture.

We evaluated the proposed method on five different real-world datasets in various application domains and demonstrated the effectiveness of GMLP compared to state-of-the-art methods in the literature. Furthermore, we conducted ablation studies and comparisons to study different architectural and training factors as well as visualizations on MNIST and synthesized data.
\ifdefined\preprint
\else
To help to reproduce the results and encouraging future studies on group-connected architectures, we made the source code related to this paper available online \footnote{We plan to include a link to the source code and GitHub page related to this paper in the camera-ready version.}.
\fi
Additional details and experimental results are provided as appendices to this paper.

\section{Related Work}
Fully-connected MLPs are the most widely-used neural models for datasets in which no prior assumption is made on the relationship between features. However, due to the huge complexity of fully-connected layers, MLPs are prone to overfitting resulting in shallow architectures limited to a few layers in depth~\citep{goodfellow2016deep}.
Various techniques have been suggested to improve training these models which include regularization techniques such as L-1/L-2 regularization, dropout, etc. and normalization techniques such as layer normalization, weigh normalization, batch normalization, etc.\citep{srivastava2014dropout,ba2016layer,salimans2016weight,ioffe2015batch}.
For instance, self-normalizing neural networks (SNNs) have been recently suggested as state of the art normalization methods that prevent vanishing or exploding gradients which help training feed-forward networks with higher depths~\citep{klambauer2017self}.

From the architectural perspective, there has been great attention toward networks consisting of sparse connections between layers rather than having dense fully-connected layers~\citep{dey2018characterizing}. Sparse connected neural networks are usually trained based on either a sparse prior structure over the network architecture~\citep{richter2018treeconnect} or based on pruning a fully-connected network to a sparse network~\citep{yun2019trimming,tartaglione2018learning,mocanu2018scalable}. However, it should be noted that the main objective of most sparse neural network literature has been focused on improving the memory and compute requirements while maintaining competitive accuracies compared to MLPs.

As a parallel line of research, the idea of using expressive feature combinations or groups has been suggested as a prior over the feature domain. Perhaps, the most successful and widespread use of this idea is in creating random forest models in which different trees are trained based on different feature subsets in order to deal with high-dimensional and high-variance data~\citep{breiman2001random}. More recently, feature grouping is suggested by \citet{aydore2019feature} as a statistical regularization technique to learn from datasets of large feature size and a small number of training samples. They do the forward network computation by projecting input features using samples taken from a bank of feature grouping matrices, reducing the input layer complexity and regularizing the model. In another recent study, \citet{ke2018tabnn} used expressive feature combinations to learn from tabular datasets using a recursive encoder with a shared embedding network. They suggest a recursive architecture in which more important feature groups have a more direct impact on the final prediction.

While promising results have been reported using these methods, feature grouping has been mostly considered as a preprocessing step. For instance, \citet{aydore2019feature} uses the recursive nearest agglomeration (ReNA)~\citep{hoyos2018recursive} clustering to determine feature groups prior to the analysis. Alternatively, \citet{ke2018tabnn} defined feature groups based on a pre-trained gradient boosting decision tree (GBDT)~\citep{friedman2001greedy}. Feature grouping as a preprocessing step not only increases the complexity and raises practical considerations, but also limits the optimality of the selected features in subsequent analysis. In this study, we propose an end-to-end solution to learn expressive feature groups. Moreover, we introduce a network architecture to exploit interrelations within the feature groups to reduce the network complexity and to train deeper representations.

\section{Proposed Method}

\begin{figure*}[ht]
    \centering
        \includegraphics[width=0.7\linewidth]{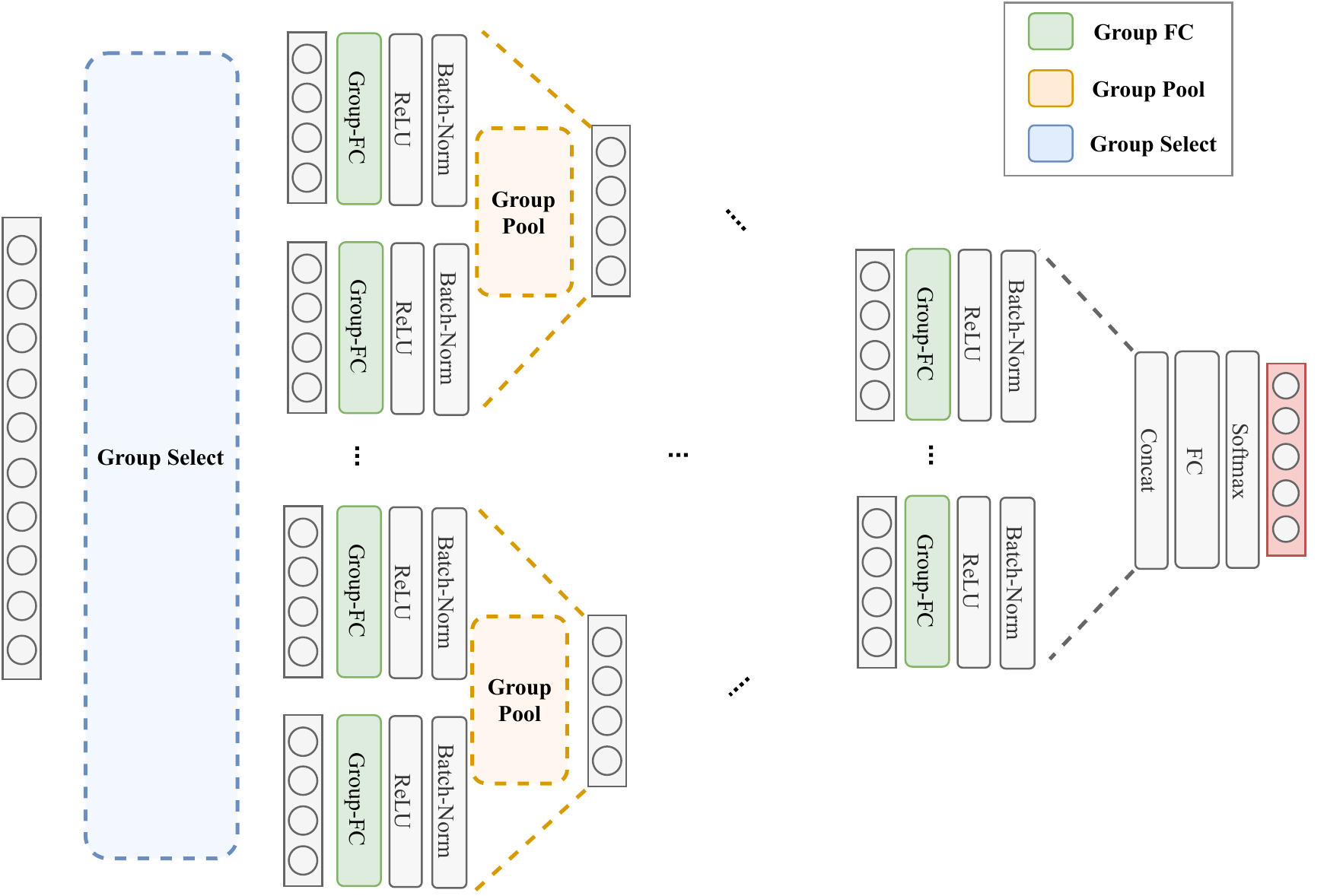}
        \caption{The GMLP network architecture.}
        \label{fig:gmlp_arch}
\end{figure*}

\subsection{Architecture Overview}
In this paper, we propose GMLP which intuitively can be broken down to three stages: $(i)$ selecting expressive feature groups, $(ii)$ learning dynamics within each group individually, and $(iii)$ merging information between groups as the network grows in depth (see Figure~\ref{fig:gmlp_arch}). In this architecture, expressive groups are jointly selected during the training phase. Furthermore, GMLP is leveraging feature groups and using local group-wise weight layers to significantly reduce the number of parameters. While the suggested idea can be materialized as different architectures, in the current study, we suggest organization of the network as architectures resembling a binary tree spanning from leaves (i.e., features) to a certain abstraction depth closer to the root\footnote{Please note that, in this paper, tree structures are considered to grow from leaves to the root . In other words, in this context, limiting the depth is synonymous with considering the tree portion spanning from a certain depth to leave nodes.}. As the network grows deeper, after each local group-wise weight layer, half of the groups are merged using pooling operations, effectively reducing the width of the network while increasing the receptive field. At the last layer, all features within all groups are concatenated into a dense feature vector fed to the output layer.

\subsection{Notation}
We consider the generic problem of supervised classification based on a dataset of feature and target pairs, $\mathcal{D}$: $(\bm{x}_{1:N},y_{1:N})$, where  $\bm{x}_i \in \Re^{d}$, $y_i \in \{1 \dots C\}$, and $N$ is the number of dataset samples. Furthermore, we define group size, $m$, as the number of neurons or elements within each group, and group count, $k$, as the number of selected groups which are essentially subsets of input features. Also, $L$ is used to refer to the total depth of a network. We use $\bm{z}^l_i \in \Re^{m}$ to refer to activation values of group $i$ in layer $l$. In this paper, we define all vectors as column vectors.

\subsection{Network Layers}
In this section, we present the formal definition of different GMLP network layers. The very first layer of the network, \texttt{Group-Select}, is responsible for organizing features into $k$ groups of size $m$ each. A routing matrix, $\Psi$, is used for connecting each neuron within each group to exactly one feature in the feature set:
\begin{equation}
    \bm{z}^0_{1:k} = \Psi \bm{x}, 
\end{equation}
where $\Psi \in \{0,1\}^{km \times d}$ is a sparse matrix determining features that are present in each group. Note that this formulation allows each feature to contribute to multiple groups as it allows features to be selected multiple times in different groups. As we are interested in jointly learning $\Psi$ during the training phase, we use the following continuous relaxation:
\begin{equation}
    \Psi_{i,j} \approx \frac{\text{exp}(\psi_{i,j} / \tau)}{\sum_{j'=1}^{j'=d} \text{exp}(\psi_{i,j'} / \tau)}.
    \label{eq:relax}
\end{equation}

In this equation, $\psi$ is a real-valued matrix reparameterizing the routing matrix through a softmax operation with temperature, $\tau$. The lower the temperature, the more \eqref{eq:relax} converges to the desired discrete and sparse binary routing matrix. Note that, in the continuous relaxation, the matrix $\psi$ can be optimized via the backpropagation of classification loss gradients. In the next section, we provide further detail on temperature annealing schedules as well as other techniques to enhance the $\Psi$ approximation.

Based on selected groups, we suggest local fully-connected weight layers for each group: \texttt{Group-FC}. The goal of \texttt{Group-FC} is to extract higher-level representations using the selected expressive feature subsets. This operation is usually followed by non-linearity functions (e.g., ReLU), normalization operations (e.g, Batch Norm), and dropout. Formally, \texttt{Group-FC} can be defined as:
\begin{equation}
    \bm{z}_{i}^{l+1} = f(W_{i}^{l} \bm{z}_{i}^{l}+\bm{b}_{i}^{l}), 
\end{equation}
where $W_{i}^{l} \in \Re^{m \times m}$ and $\bm{b}_{i}^{l} \in \Re^{m}$ are the weight matrix and bias vector, applied on group $i$ at layer $l$. Here, $f$ represents other subsequent operations such as non-linearity, normalization, and dropout.

Lastly, \texttt{Group-Pool} is defined as an operation which merges representations of two groups into a single group, reducing network width by half while increasing the effective receptive field:
\begin{equation}
     \bm{z}_{i}^{l+1} = pool(\bm{z}_{i}^{l},\bm{z}_{i+k/{2^{l+1}}}^{l}),
\end{equation}
where $\bm{z}_{i}^{l}$ and $\bm{z}_{i+k/2}^{l}$ are the $i$th group from the first and second halves, respectively; and $pool$ is a pooling function from $\Re^{2m}$ to $\Re^{m}$. In this study, we explore different variants of pooling functions such as max pooling, average pooling, or using linear weight layers as transformations from $\Re^{2m}$ to $\Re^{m}$. Please note that while we use a similar terminology as pooling in convolutional networks, the pooling operation explained here is not applied location-wise, but instead it is applied feature-wise, between different groups pairs.

The values of $m$ and $k$ are closely related to the number and order of feature interactions for a certain task. Using proper $m$ and $k$ values enables us to reduce the parameter space while maintaining the model complexity required to solve the task. However, finding the ideal $m$ and $k$ directly from a given dataset is a very challenging problem. In this work, we treat $m$ and $k$ as hyperparameters to be found by a hyperparameter search.


\subsection{Training}
We define the objective function to be used for end-to-end training of weights as well as the routing matrix as:
\begin{equation}
    L = - \frac{1}{N} \sum_i \sum_{c} y_{i,c} \; \text{log}(F_{\theta}(\bm{x}_i))  + \lambda H(\psi) + \alpha \sum_{\omega \in \theta} ||\omega||_2^{2}. 
    \label{eq:obj}
\end{equation}
In this objective function, the first term is the standard cross-entropy classification loss where $F_{\theta}$ denotes the GMLP network as a function with parameters $\theta$, and $N$ is the number of training samples used. The second term is an entropy loss over the distribution of the routing matrix that is weighted by the hyperparameter $\lambda$:

\begin{equation}
    H(\psi) = -\frac{1}{d} \sum_{j=1}^{j=d} \sum_{i=1}^{i=km}  \frac{\text{exp}(\psi_{i,j})}{\sum_{j'=1}^{j'=d} \text{exp}(\psi_{i,j'})} \text{log}(\frac{\text{exp}(\psi_{i,j})}{\sum_{j'=1}^{j'=d} \text{exp}(\psi_{i,j'})}).
\end{equation}

$H(\psi)$ is minimizing the entropy corresponding to the distribution of $\psi$ regardless of the temperature used for $\Psi$ approximation. Accordingly, $\lambda$ can be viewed as a hyperparameter and as an additional method for encouraging sparse $\Psi$ matrices. The last term in \eqref{eq:obj} is an L-2 regularization term with the hyperparameter $\alpha$ to control the magnitude of parameters in layer weights and in $\psi$. Note that without the L-2 regularization term, $\psi$ elements may keep increasing during the optimization loop, since $\psi$ only appears in normalized form in the objective function of \eqref{eq:obj}.

We use Adam~\citep{kingma2014adam} optimization algorithm starting from the default 0.001 learning rate and reducing the learning rate by a factor of 5 as the validation accuracy stops improving. Regarding the temperature annealing, during the training, the temperature is exponentially decayed from 1.0 to 0.01. In order to initialize the \texttt{Group-FC} weights, we used Xavier initialization~\citep{glorot2010understanding} with \texttt{m} for both fan-in and fan-out values. Similarly, the $\psi$ matrix is initialized by setting the fan-in equal to $d$ and fan-out to $km$. 

Further detail on architectures and hyperparameters used for each specific experiment as well as details on the software implementation are provided as appendices to this paper.

\subsection{Analysis}
\label{sec:Analysis}
The computational complexity of GMLP at the prediction time can be written as (for simplicity, ignoring bias and pooling terms):
\begin{equation}
    km + km^2 + \frac{km^2}{2} + \frac{km^2}{4} + ... + \frac{km^2}{2^{L-1}} + C\frac{km}{2^{L-1}}.
\end{equation}
In this series, the first term, $km$, is the work required to organize features to groups. The subsequent terms, except the last term, are representing the computational cost of local fully-connected operations at each layer. The last term is the complexity of the output layer transformation from the concatenated features to the number of classes. Therefore, the computational complexity of GMLP at the prediction time can be written as $\mathcal{O}(km^2+\frac{Ckm}{2^{L-1}})$. In comparison, the computational complexity of an MLP with a similar network width would be:
\begin{equation}
    kmd + k^2m^2 + \frac{k^2m^2}{2} + \frac{k^2m^2}{4} + ... + \frac{k^2m^2}{2^{L-1}} + C\frac{km}{2^{L-1}},
\end{equation}
where the first term is the work required for the first network layer from $d$ to $km$ neurons, the second term is corresponding to a hidden layer of size $km$, and so forth. The last term is the complexity of the output layer similar to the case of GMLP. The overall work required from this equation is of $\mathcal{O}(kmd+k^2m^2+\frac{Ckm}{2^{L-1}})$. This is substantially higher than GMLP, for typical $k$, $d$, and $C$ values.

Additionally, the density of the \texttt{Group-FC} layer connections can be calculated as: $\frac{km^2}{k^2m^2} = \frac{1}{k}$, which is very small for reasonably large number of $k$ values used in our experiments. Also, assuming pooling operations in every other layer, the receptive field size or the maximum number of features impacting a neuron at layer $l$ can be written as $2^{l-1}m$. For instance, a neuron in the first layer of the network is only connected to $m$ features, and a neuron in the second layer is connected to two groups or $2m$ features and so forth.

\section{Experiments}
\label{sec:Experiments}
\subsection{Experimental Setup}
\begin{table*}[h]%
\renewcommand{\arraystretch}{1.4}
\caption{Summary of datasets used in our experiments.}
\begin{minipage}{\textwidth}
\begin{center}
\resizebox{\textwidth}{!}{
\begin{tabular}{l|cccccc}
\toprule

\textbf{Dataset} & \textbf{\# Train Samples} & \textbf{\# Test Samples} & \textbf{\# Features} & \textbf{\# Classes} & \textbf{Domain}\\

\hline

\textbf{CIFAR-10\footnote{Permuted version, i.e. pixel coordinates are ignored.}}~{\small \citep{krizhevsky2009learning}} & 50,000 & 10,000 & 3,072 & 10 & Image Classification \\

\textbf{HAPT}~{\small \citep{anguita2013public}} & 6,002 & 2,451 & 561 & 5 & Activity Recognition \\

\textbf{Tox21\footnote{Aryl hydrocarbon Receptor (AhR) activity prediction task adapted from \citet{mayr2016deeptox}.}}~{\small \citep{huang2016tox21challenge}} & 8,441 & 610 & 1,644 & 2 & Drug Discovery \\


\textbf{Diabetes\footnote{Data processing pipeline adopted from \citet{kachuee2019nutrition}}}~{\small (This Work)} & 47,125 & 11,782 & 116 & 2 & Disease Diagnosis \\

\textbf{Hypertension$^c$}~{\small (This Work)} & 49,819 & 12,455 & 121 & 2 & Disease Diagnosis \\

\textbf{Cholesterol$^c$}~{\small (This Work)} & 54,360 & 13,591 & 120 & 2 & Disease Diagnosis \\

\textbf{Landsat}~{\small \citep{Dua2019}} & 4,435 & 2,000 & 36 & 6 & Satellite Imaging \\

\textbf{MIT-BIH\footnote{We use preprocessed data from {http://kaggle.com/shayanfazeli/heartbeat}}}~{\small \citep{moody2001impact}} & 87,554 & 21,892 & 187 & 5 & ECG Classification \\

\textbf{MNIST}~{\small \citep{lecun2010mnist}} & 60,000 & 10,000 & 784 & 10 & Digit Classification \\

\textbf{Synthesized}~{\small (Appendix~D)} & 5,120 & 1,280 & 6 & 2 & See Appendix~D \\

\bottomrule
\end{tabular}
}
\end{center}
\end{minipage}
\label{tab:datasets}
\end{table*}%

The proposed method is evaluated on five different real-world datasets, covering various domains and applications: permutation invariant CIFAR-10~\citep{krizhevsky2009learning}, human activity recognition (HAPT)~\citep{anguita2013public}, toxicity prediction (Tox21)~\citep{huang2016tox21challenge}, and UCI Landsat~\citep{Dua2019}, and MIT-BIH arrhythmia classification~\citep{moody2001impact}.
Additionally, we use three real-world tabular datasets in health domain: diabetes, hypertension, cholesterol classification tasks~\citep{kachuee2019nutrition}. 
We use MNIST~\citep{lecun2010mnist} and a synthesized dataset to provide further insight into the operation of GMLP (see Appendix). Table~\ref{tab:datasets} presents a summary of datasets used in this study. Regarding the CIFAR-10 dataset, we permute the image pixels to discard pixel coordinates in our experiments. Note that the permutation is not changing across samples, it is merely a fixed random ordering used to remove pixel coordinates for each experiment. For all datasets, basic statistical normalization with $\mu=0$ and $\sigma=1$ is used to normalize features as a preprocessing step. The only exception is CIFAR-10 for which we used the standard channel-wise normalization and standard data augmentation (i.e., random crops and random horizontal flips). The standard test and train data splits were used as dictated by dataset publishers. In cases that the separated sets are not provided, test and train subsets are created by randomly splitting samples to $20\%$ for test and the rest for training/validation.

We compare the performance of the proposed method with recent related work including Self-Normalizing Neural Networks (SNN)~\citep{klambauer2017self}, Sparse Evolutionary Training (SET)~\citep{mocanu2018scalable}\footnote{https://github.com/dcmocanu/sparse-evolutionary-artificial-neural-networks}, Feature Grouping as a Stochastic Regularizer (in this paper, denoted as FGR)~\citep{aydore2019feature}\footnote{https://github.com/sergulaydore/Feature-Grouping-Regularizer} as well as the basic dropout regularized and batch normalized MLPs. Additionally, as a non-neural baseline, we make comparisons with random forest classifiers (RFC) consisting of $1000$ trees trained using the gini criterion~\citep{liaw2002classification}. In order to ensure a fair comparison, we adapted source codes provided by other work to be compatible with our data loader and preprocessing modules.


Furthermore, for each method, we conducted an extensive hyperparameter search using Microsoft Neural Network Intelligence (NNI) toolkit\footnote{https://github.com/microsoft/nni} and the Tree-structured Parzen Estimator (TPE) tuner~\citep{bergstra2011algorithms} covering different architectural and learning hyperparameters for each case. More detail on hyperparameter search spaces and specific architectures used in this paper is provided in the appendices. We run each case using the best hyperparameter configuration eight times and report mean and standard deviation values.

\subsection{Results}
Table~\ref{tab:accs} presents a comparison between the proposed method (GMLP) and 5 other baselines: MLP, SNN~\citep{klambauer2017self}, SET~\citep{mocanu2018scalable}, FGR~\citep{aydore2019feature}, and RFC. From this comparison, GMLP outperforms other work, achieving state-of-the-art classification accuracies. Concerning the CIFAR-10 results, to the best of our knowledge, GMLP achieves a new state-of-the-art performance on permutation invariant CIFAR-10 augmented using the standard data augmentation. Note that \citet{lin2015far} reported 78\% accuracy on the permuted CIFAR-10 using additional non-standard augmentations and about 70\% otherwise. We believe that leveraging expressive feature groups enables GMLP to consistently perform better across different datasets.

\begin{table}[h]%
\renewcommand{\arraystretch}{1.4}
\caption{Comparison of top-1 test accuracies for GMLP and other work.}
\begin{minipage}{1.0\textwidth}
\begin{center}
\resizebox{\columnwidth}{!}{
\begin{tabular}{l|cccccc}
\toprule

 & \multicolumn{6}{c}{\textbf{Accuracy (\%)}} \\

\textbf{Dataset} & \textbf{GMLP} & \textbf{MLP} & \textbf{SNN\footnote{Self-Normalizing Neural Networks \citep{klambauer2017self}}} & \textbf{SET\footnote{Sparse Evolutionary Training \citep{mocanu2018scalable}}} & \textbf{FGR\footnote{Feature Grouping as a Stochastic Regularizer \citep{aydore2019feature}}} & \textbf{RFC}  \\


\hline

\textbf{CIFAR-10
}~{\small \citep{krizhevsky2009learning}} & \textbf{73.76} {\scriptsize($\pm 0.14$)} & 68.15 {\scriptsize($\pm 0.56$)} & 66.88 {\scriptsize($\pm 0.30$)} & 72.71 {\scriptsize($\pm 0.29$)} & 45.90 {\scriptsize($\pm 0.40$)} & -\\

\textbf{HAPT}~{\small \citep{anguita2013public}} & \textbf{96.34} {\scriptsize($\pm 0.19$)} & 95.73 {\scriptsize($\pm 0.38$)} & 95.47 {\scriptsize($\pm 0.09$)} & 71.35 {\scriptsize($\pm 0.74$)} & 91.57 {\scriptsize($\pm 0.31$)} & 91.51 {\scriptsize($\pm 0.10$)}\\

\textbf{Tox21\footnote{Percentage of the area under the ROC curve is reported for this dataset.}}~{\small \citep{huang2016tox21challenge}} & \textbf{88.57} {\scriptsize($\pm 0.36$)} & 86.90 {\scriptsize($\pm 0.38$)} & 86.68 {\scriptsize($\pm 0.88$)} & 88.29 {\scriptsize($\pm 0.20$)} & 87.17 {\scriptsize($\pm 0.38$)} & 88.42 {\scriptsize($\pm 0.19$)}\\


\textbf{Diabetes$^d$}~{\small (This Work)} & \textbf{88.37} {\scriptsize($\pm 0.04$)} & 87.80 {\scriptsize($\pm 0.03$)} & 87.83 {\scriptsize($\pm 0.02$)} & 72.39 {\scriptsize($\pm 2.32$)} & 87.31 {\scriptsize($\pm 0.05$)} & 87.51 {\scriptsize($\pm 0.15$)}\\

\textbf{Hypertension$^d$}~{\small (This Work)} & \textbf{87.26} {\scriptsize($\pm 0.04$)} & 86.83 {\scriptsize($\pm 0.05$)} & 86.93 {\scriptsize($\pm 0.01$)} & 82.92 {\scriptsize($\pm 1.93$)} & 86.29 {\scriptsize($\pm 0.05$)} & 86.24 {\scriptsize($\pm 0.06$)}\\

\textbf{Cholesterol$^d$}~{\small (This Work)} & \textbf{83.18} {\scriptsize($\pm 0.06$)} & 82.66 {\scriptsize($\pm 0.05$)} & 82.62 {\scriptsize($\pm 0.10$)} & 76.59 {\scriptsize($\pm 2.15$)} & 82.10 {\scriptsize($\pm 0.07$)} & 82.48 {\scriptsize($\pm 0.04$)}\\

\textbf{Landsat}~{\small \citep{Dua2019}} & \textbf{91.54} {\scriptsize($\pm 0.16$)} & 91.21 {\scriptsize($\pm 0.44$)} & 91.37 {\scriptsize($\pm 0.21$)} & 91.03 {\scriptsize($\pm 0.56$)} & 90.70 {\scriptsize($\pm 0.17$)} & 91.15 {\scriptsize($\pm 0.09$)}\\

\textbf{MIT-BIH}~{\small \citep{moody2001impact}} & \textbf{98.74} {\scriptsize($\pm 0.04$)} & 98.65 {\scriptsize($\pm 0.01$)} & 98.56 {\scriptsize($\pm 0.02$)} & 98.10 {\scriptsize($\pm 0.01$)} & 98.13 {\scriptsize($\pm 0.03$)} & 98 .46 {\scriptsize($\pm 0.01$)}\\

\bottomrule
\end{tabular}
}
\end{center}
\end{minipage}
\label{tab:accs}
\end{table}%

\begin{figure}[h]
    \centering
    \begin{subfigure}[b]{0.42\columnwidth}
        \centering
        \includegraphics[width=\columnwidth]{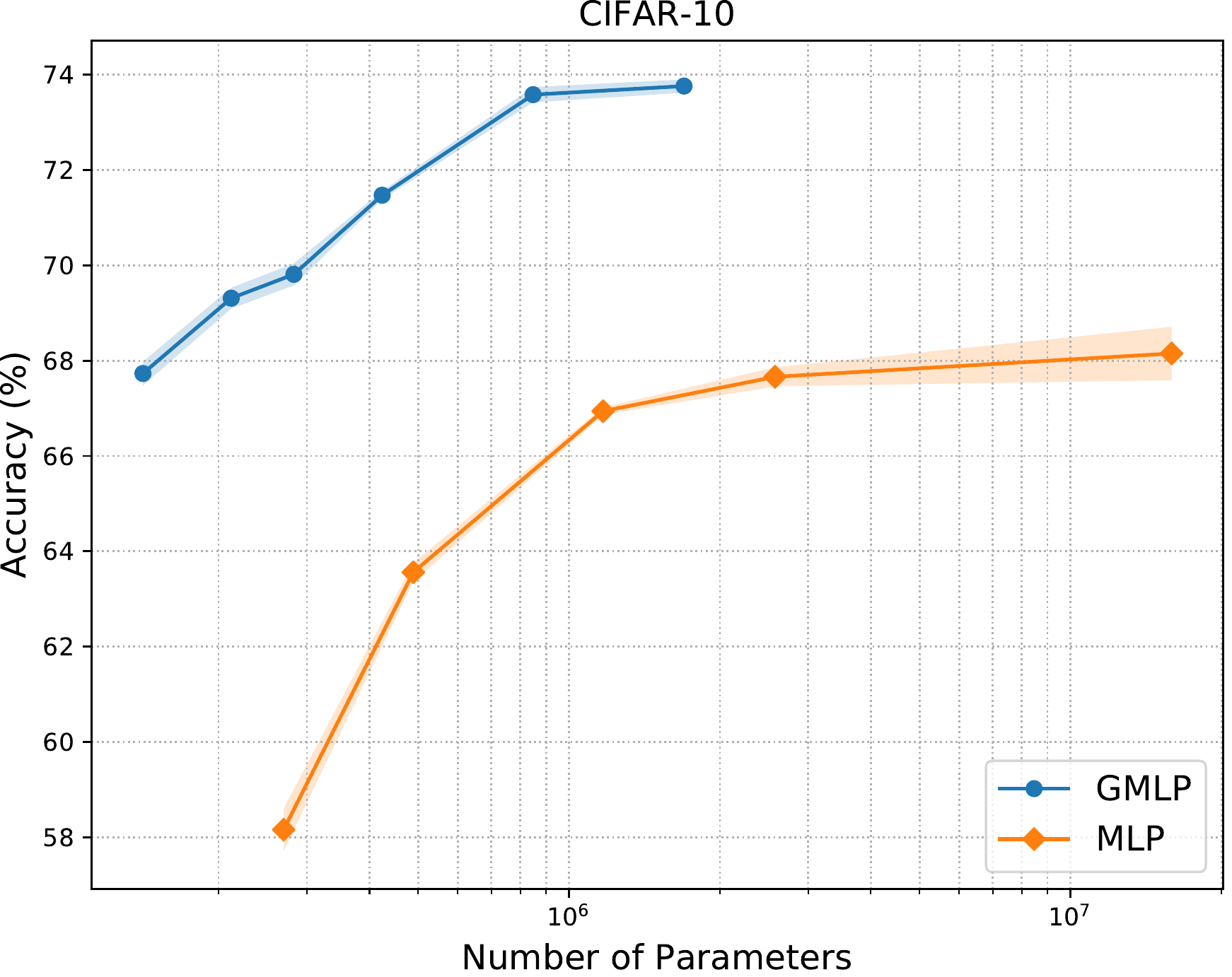}
        \caption{}
        \label{fig:params_cifar10_2}
    \end{subfigure}
    ~
    \begin{subfigure}[b]{0.42\columnwidth}
        \centering
        \includegraphics[width=\columnwidth]{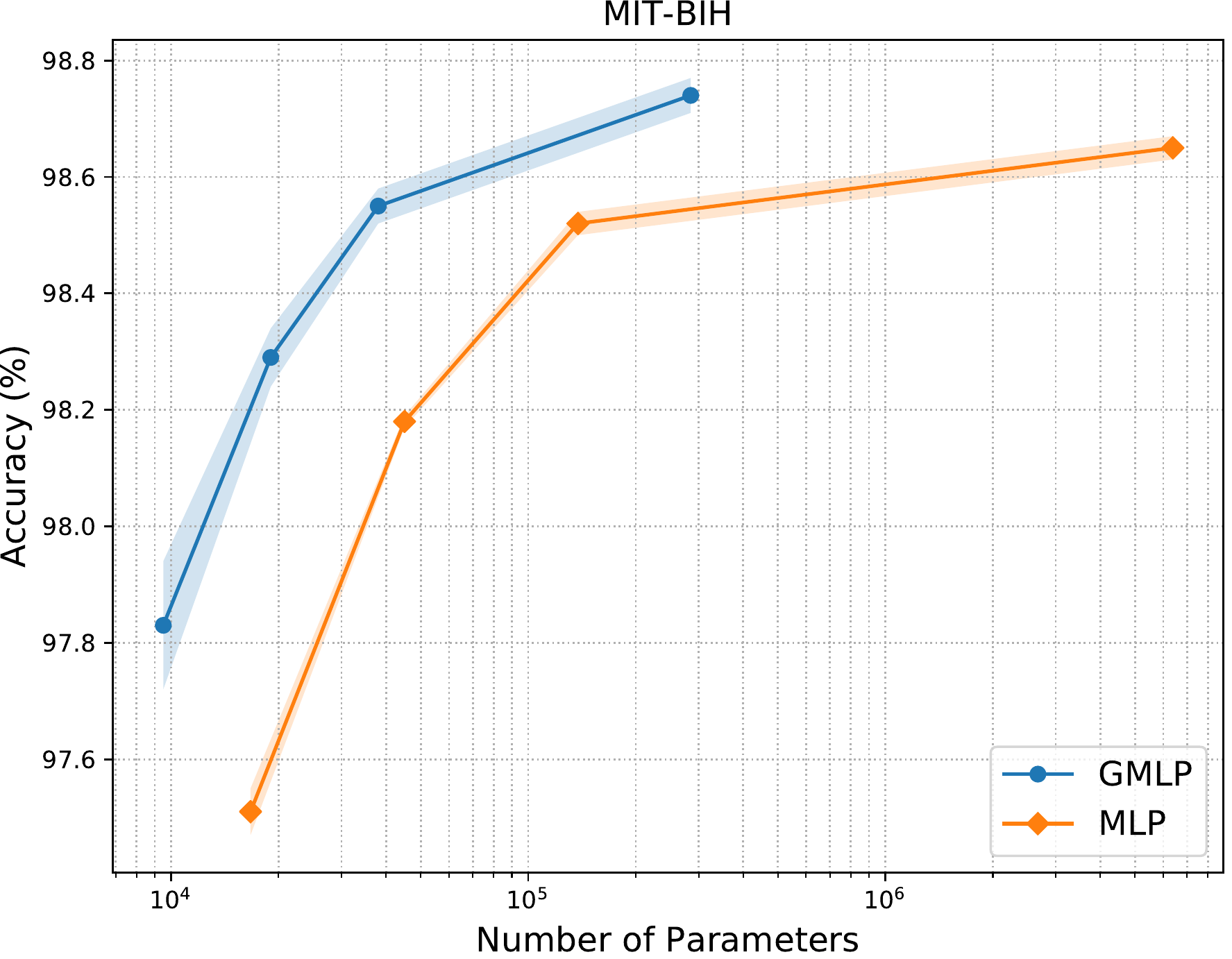}
        \caption{}
        \label{fig:params_mitbih}
    \end{subfigure}
    \caption{Accuracy versus number of parameters for this work (GMLP) and the MLP baseline: (a) CIFAR-10 dataset, (b) MIT-BIH dataset. The x-axis is in a logarithmic scale.}
    \label{fig:params}
    \vspace{-0.0in}
\end{figure}

To compare model complexity and performance we conduct an experiment by changing the number of model parameters and reporting the resulting test accuracies. Here, we reduce the number of parameters by reducing the width of each network; i.e. reducing the number of groups and hidden neurons for GMLP and MLP, respectively. Figure~\ref{fig:params} shows accuracy versus the number of parameters for the GMLP and MLP baseline on CIFAR-10 and MIT-BIH datasets. Based on this figure, GMLP is able to achieve higher accuracies using significantly less number of parameters. It is consistent with the complexity analysis provided in Section~\ref{sec:Analysis}. Note that in this comparison, we consider the number of parameters involved at the prediction time.

\subsection{Ablation Study}
\begin{figure}[t]
\centering
\begin{minipage}[t]{0.45\columnwidth}
    \centering
        \includegraphics[width=\columnwidth]{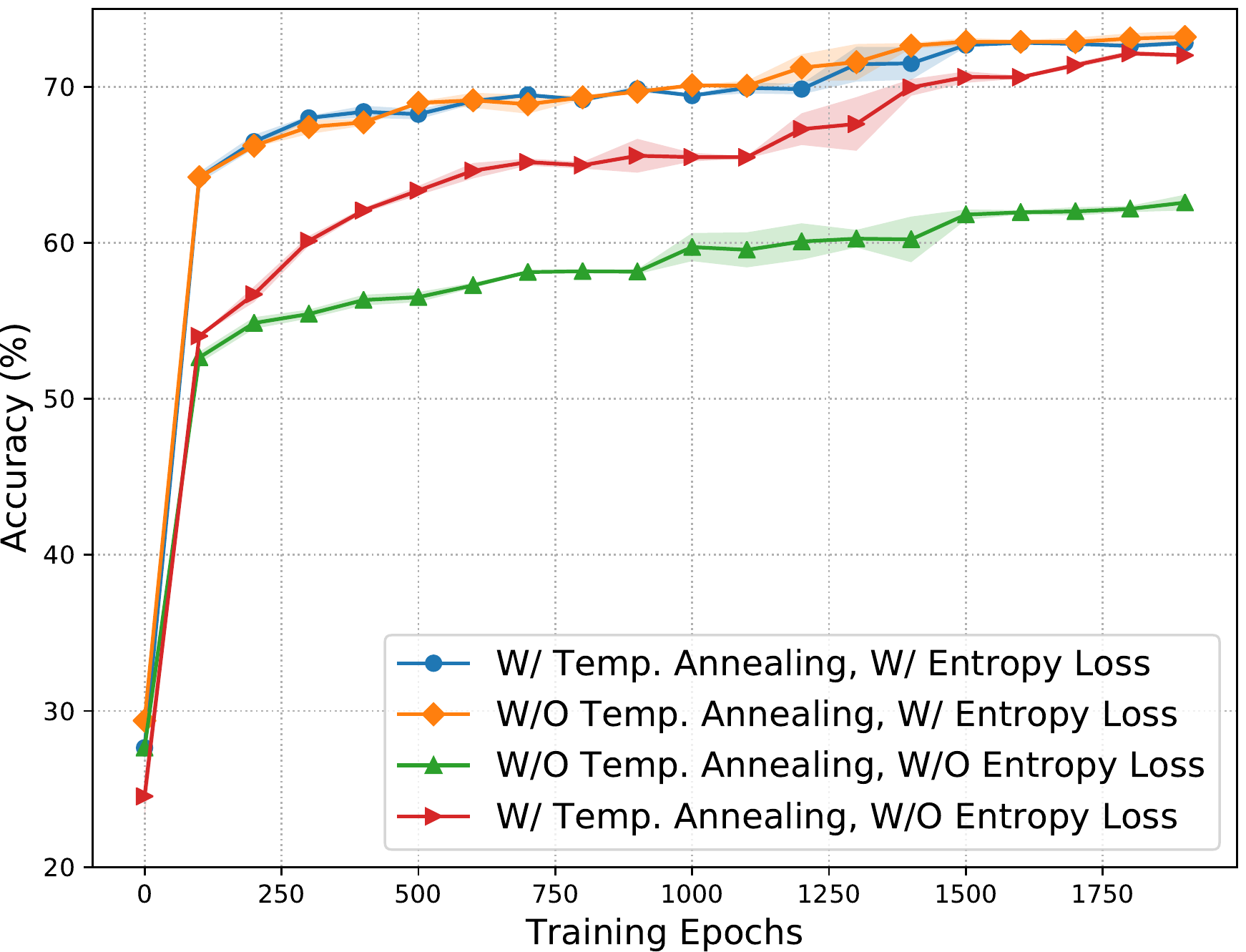}
        \caption{Ablation study on the impact of temperature annealing and entropy loss terms.}
        \label{fig:ablation_cifar10_temphloss}
\end{minipage}
~
\begin{minipage}[t]{0.45\columnwidth}
    \centering
        \includegraphics[width=\columnwidth]{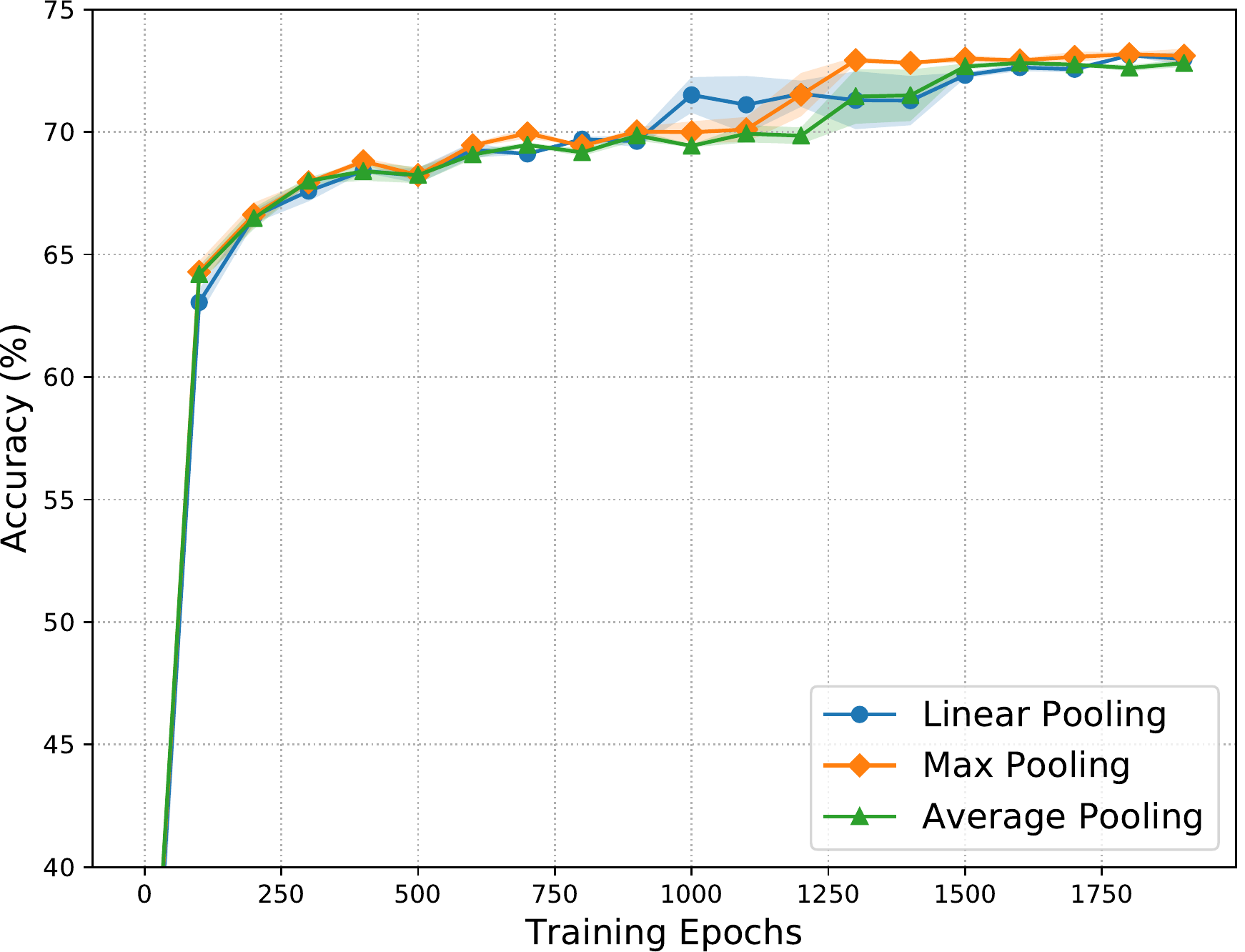}
        \caption{Ablation study demonstrating the impact of different pooling functions.}
        \label{fig:ablation_cifar10_pooling}
\end{minipage}
\vspace{-0.1in}
\end{figure}

Figure~\ref{fig:ablation_cifar10_temphloss} presents an ablation study comparing the performance of GMLP on CIFAR-10 dataset for networks trained: $(i)$  using both the temperature annealing and the entropy loss objective, $(ii)$ using only temperature annealing without the entropy loss objective, $(iii)$ using no temperature annealing but using the entropy loss objective, $(iv)$ not using any of the temperature annealing or the entropy loss objective.
From this figure, excluding both techniques leads to a significantly lower performance. However, using any of the two techniques leads to relatively similar high accuracies. It is consistent with the intuition that the functionality of these techniques is to encourage learning sparse routing matrices, either using softmax temperatures or entropy regularization to achieve this. In this paper, in order to ensure sparse routing matrices, we use both techniques simultaneously as in case $(i)$.

Figure~\ref{fig:ablation_cifar10_pooling} shows a comparison between GMLP models trained on CIFAR-10 using different pooling types: $(i)$ linear transformation, $(ii)$ max pooling, and $(iii)$ average pooling. As it can be seen from this comparison, while there are slight differences in the convergence speed of using different pooling types, all of them achieve relatively similar accuracies. In our experiments, we decided to use max pooling and average pooling as they provide reasonable results without the need to introduce additional parameters required for the linear pooling method.

\begin{figure}[t]
\centering
\begin{minipage}[t]{0.42\columnwidth}
    \centering
        \includegraphics[width=\columnwidth]{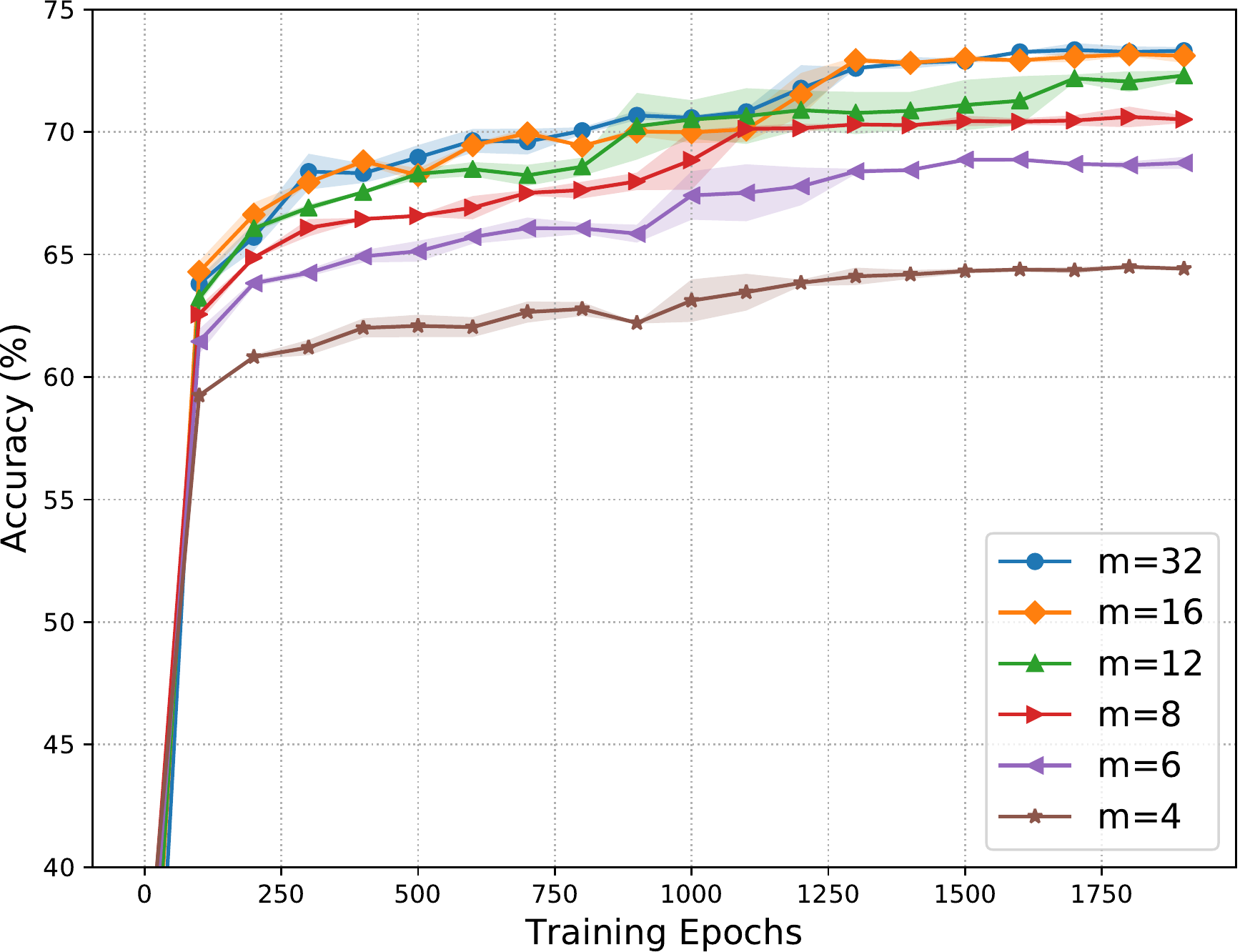}
        \caption{Ablation study on the impact of using different group sizes ($m$). For this experiment, we used $k$=1536. }
        \label{fig:ablation_cifar10_m}
\end{minipage}
~
\begin{minipage}[t]{0.42\columnwidth}
    \centering
        \includegraphics[width=\columnwidth]{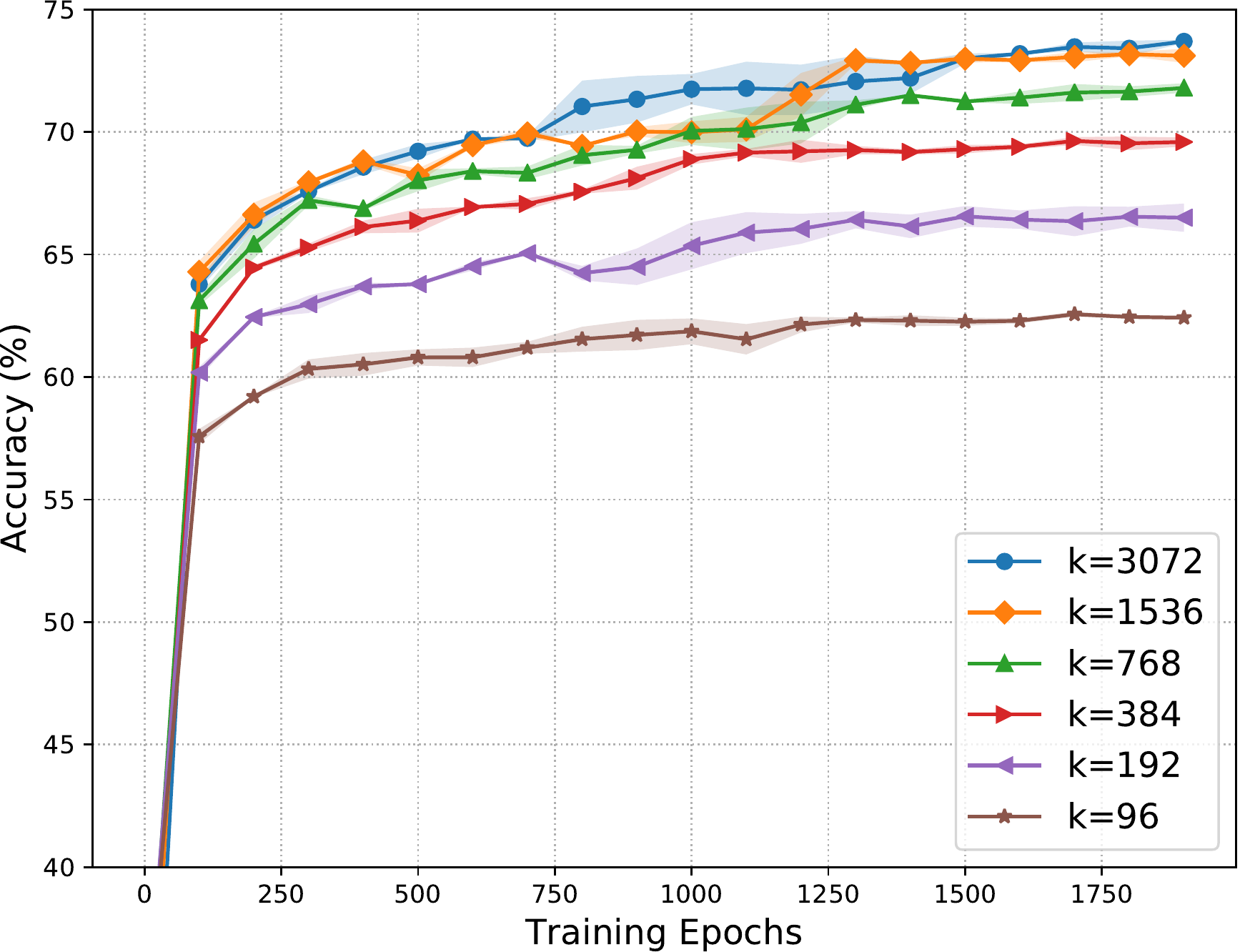}
        \caption{Ablation study on the impact of using different number of groups ($k$). For this experiment, we used $m$=16.}
        \label{fig:ablation_cifar10_k}
\end{minipage}
\vspace{-0.1in}
\end{figure}

Figure~\ref{fig:ablation_cifar10_m} shows learning curves for training CIFAR-10 models using different group sizes. From this figure, using very small group sizes would cause a reduction in the final accuracy. At the other extreme, the improvement achieved using larger values is negligible for $m$ values more than 16. Finally, Figure~\ref{fig:ablation_cifar10_k} shows a comparison between learning curves for using a different number of groups. Using very small $k$ values result in a significant reduction in performance. However, the rate of performance gains for using more groups is very small for $k$ of more than 1536. Note that the number of model parameters and compute scales linearly with $k$ and quadratically with $m$ (see Section~\ref{sec:Analysis}).

\section{Discussion}
Intuitively, training a GMLP model with certain groups can be viewed as a prior assumption over the number and order of interactions between the features. It is a reasonable prior assumption as in many natural datasets, a conceptual hierarchy exists where only a limited number of features interact with each other. 
Additionally, GMLP can be considered as a more general neural counterpart of random forests. Both models use subsets of features (i.e., groups) and learn interactions within each group. A major difference between the two methods is that GMLP combines information between different groups using pooling operations, while random forest trains an ensemble of independent trees on each group. From another perspective, the idea of studying feature groups is related to causal models such as Bayesian networks and factor graphs \citep{darwiche2009modeling,neapolitan2004learning,clifford1990markov}. These methods are often impractical for large-scale problems, because without a prior over the causal graph, they require an architecture search of the NP-complete complexity or more.

\section{Conclusion}
In this paper, we proposed GMLP as a solution for deep learning in domains where the feature interactions are not known as prior and do not admit the use of convolutional or other techniques leveraging domain priors. GMLP jointly learns expressive feature combinations and employs group-wise operations to reduce the network complexity.  We conducted extensive experiments demonstrating the effectiveness of the proposed idea compared to the state-of-the-art methods in the literature.


\FloatBarrier
\bibliographystyle{template/iclr2021/iclr2021_conference}
\bibliography{bib.bib}

\clearpage
\appendix

\section{Hyperparameter Search Space}
\label{sec:search}
Tables~\ref{tab:search_gmlp}-\ref{tab:search_fgr} present the hyperparameter search space considered for experiments on GMLP, MLP, SNN, and FGR, respectively. For the GMLP search space, the number of groups is adjusted based on the number of features and samples in each specific task. Also, the number of layers is adjusted to be compatible with the number of groups being used. Regarding the FGR experiments, due to scalability issues of the published source provided by the original authors, we were only able to train networks with at most two hidden layers. For SET, as their architecture is evolutionary i.e., prunes certain weights and adds new ones, we only explored using a different number of hidden neurons in the range of 500 to 4000. Regarding the RFC baseline, we found that the Gini criterion tends to work very similar or slightly better than the entropy measure. Also, for our experiments, using an ensemble of 1000 trees was more than sufficient to train stable and powerful RFC models. However, we found that the maximum depth of trees is an important hyperparameter to be adjusted for each experiment.

Regarding the number of epochs, we used 2000 epochs for CIFAR-10, 1000 epochs for HAPT, and 300 epochs for the rest of the datasets. The only exception is the SNN experiments where we had to reduce the learning rate to increase the stability of the training resulting in more epochs required to converge.

\begin{table}[h]%
\renewcommand{\arraystretch}{1.4}
\caption{Hyperparameter search space used for GMLP experiments.}
\begin{center}
\begin{tabular}{l|l}
\toprule
\textbf{Hyperparameter} & \textbf{Considered Values}\\
\hline
Number of hidden layers & $\{1,2,...,8\}$\footnote{The range is adjusted based on the number of groups.} \\
Number of groups & $16-4096$\footnote{The range is adjusted based on the number of features and sample size.} \\
Size of groups & $4-32$ \\
Lambda & $[10^{-2} , 10^{4}]$ \\
Alpha & $[10^{-12} , 10^{-1}]$ \\
Dropout rate & $[0,1]$ \\
\bottomrule
\end{tabular}
\end{center}
\label{tab:search_gmlp}
\end{table}%

\begin{table}[h]%
\renewcommand{\arraystretch}{1.4}
\caption{Hyperparameter search space used for MLP and SNN experiments.}
\begin{center}
\begin{tabular}{l|l}
\toprule
\textbf{Hyperparameter} & \textbf{Considered Values}\\
\hline
Number of hidden layers & $\{1,2,...,6\}$\\
Size of hidden layers & [$0.05 \times n_{features}] - [20 \times n_{features}]$\\
Alpha & $[10^{-12} , 10^{-1}]$ \\
Dropout rate & $[0,1]$ \\
\bottomrule
\end{tabular}
\end{center}
\label{tab:search_mlp_snn}
\end{table}%

\begin{table}[h]%
\renewcommand{\arraystretch}{1.4}
\caption{Hyperparameter search space used for FGR experiments.}
\begin{center}
\begin{tabular}{l|l}
\toprule
\textbf{Hyperparameter} & \textbf{Considered Values}\\
\hline
Number of hidden layers & $\{1,2\}$\\
Size of hidden layers & $n_{features} - [50 \times n_{features}]$\\
Number of groups & $2 - n_{features} $ \\
Alpha & $[10^{-12} , 10^{-1}]$ \\
\bottomrule
\end{tabular}
\end{center}
\label{tab:search_fgr}
\end{table}%

\clearpage
\section{Architectures}
\label{sec:archs}

Table~\ref{tab:archs_gmlp},\ref{tab:archs_mlp},\ref{tab:archs_snn},\ref{tab:archs_set},\ref{tab:archs_fgr} show the selected architectures for GMLP, MLP, SNN, SET, and FGR, respectively. We used the following notation to indicate different layer types and parameters: \texttt{GSel-k-m} represents a \texttt{Group-Select} layer selecting \texttt{k} groups of \texttt{m} features each. \texttt{GFC} indicates \texttt{Group-FC} layers, and \texttt{FC-x} represents fully-connected layer with \texttt{x} hidden neurons. \texttt{GPool-x} is a \texttt{Group-Pool} layer of type \texttt{x} (max, mean, linear, etc.). \texttt{Concat} is concatenation of groups used prior to the output layer in GMLP architectures. \texttt{SC-x} refers to SET sparse evolutionary layer of size \texttt{x}.

\begin{wraptable}{l}{1.0\linewidth}
\centering
\renewcommand{\arraystretch}{1.4}
\caption{GMLP architectures used in our experiments.}
\begin{center}
\resizebox{0.80\textwidth}{!}{
\begin{tabular}{l|l}
\toprule
\textbf{Dataset} & \textbf{Architecture}\\
\hline

\multirow{3}{*}{\textbf{CIFAR-10}} & 
\texttt{GSel-1536-16, GFC, ReLU, BNorm, GPool-max, GFC, ReLU, BNorm, }\\
& \texttt{GPool-max, GFC, ReLU, BNorm, GPool-max, GFC, ReLU, BNorm, }\\
& \texttt{Concat, FC-10, Softmax} \\
\hline
\multirow{2}{*}{\textbf{HAPT}} & \texttt{GSel-288-12, GFC, ReLU, BNorm, GPool-mean, GFC, ReLU, BNorm, }\\
 & \texttt{Concat, FC-5, Softmax} \\
\hline
\multirow{3}{*}{\textbf{Tox21}} & 
\texttt{GSel-320-28, GFC, ReLU, BNorm, GPool-mean, GFC, ReLU, BNorm, }\\
& \texttt{GPool-mean, GFC, ReLU, BNorm, GPool-mean, GFC, ReLU, BNorm, }\\
& \texttt{GPool-mean, GFC, ReLU, BNorm, Concat, FC-2, Softmax} \\
\hline
\multirow{3}{*}{\textbf{Diabetes}} & 
\texttt{GSel-352-4, GFC, ReLU, BNorm, GPool-mean, GFC, ReLU, BNorm, }\\
& \texttt{GPool-mean, GFC, ReLU, BNorm, GPool-mean, GFC, ReLU, BNorm, }\\
& \texttt{GPool-mean, Concat, FC-2, Softmax} \\
\hline
\multirow{2}{*}{\textbf{Hypertension}} & 
\texttt{GSel-448-8, GFC, ReLU, BNorm, GPool-mean, GFC, ReLU, BNorm, }\\
& \texttt{GPool-mean, Concat, FC-2, Softmax} \\
\hline
\multirow{3}{*}{\textbf{Cholesterol}} & 
\texttt{GSel-384-24, GFC, ReLU, BNorm, GPool-mean, GFC, ReLU, BNorm, }\\
& \texttt{GPool-mean, GFC, ReLU, BNorm, GPool-mean, GFC, ReLU, BNorm, }\\
& \texttt{GPool-mean, Concat, FC-2, Softmax} \\
\hline
\multirow{2}{*}{\textbf{Landsat}} & 
\texttt{GSel-88-16, GFC, ReLU, BNorm, GPool-mean, GFC, ReLU, BNorm, }\\
& \texttt{GPool-mean, GFC, ReLU, BNorm, GPool-mean, Concat, FC-6, Softmax}\\
\hline
\multirow{2}{*}{\textbf{MIT-BIH}} & 
\texttt{GSel-240-24, GFC, ReLU, BNorm, GPool-mean, GFC, ReLU, BNorm, }\\
& \texttt{GPool-mean, GFC, ReLU, BNorm, GPool-mean, Concat, FC-5, Softmax}\\
\hline
\multirow{3}{*}{\textbf{MNIST}} & 
\texttt{GSel-64-16, GFC, ReLU, BNorm, GPool-mean, GFC, ReLU, BNorm, }\\
& \texttt{GPool-mean, GFC, ReLU, BNorm, GPool-mean, GFC, ReLU, BNorm, }\\
& \texttt{Concat, FC-10, Softmax} \\
\hline

\multirow{1}{*}{\textbf{Synthesized}} & 
\texttt{GSel-4-2, GFC, ReLU, BNorm, Concat, FC-2, Softmax}\\

\bottomrule
\end{tabular}
}
\end{center}
\label{tab:archs_gmlp}
\end{wraptable}

\begin{table*}[h]%
\renewcommand{\arraystretch}{1.4}
\caption{MLP architectures used in our experiments.}
\begin{minipage}{\columnwidth}
\begin{center}
\resizebox{\columnwidth}{!}{
\begin{tabular}{l|l}
\toprule
\textbf{Dataset} & \textbf{Architecture}\\
\hline
\multirow{2}{*}{\textbf{CIFAR-10}} & 
\texttt{FC-3072, ReLU, BNorm, FC-2764, ReLU, BNorm, }\\
& \texttt{FC-2488, ReLU, BNorm, FC-10, Softmax} \\
\hline
\multirow{1}{*}{\textbf{HAPT}} & \texttt{FC-106, ReLU, BNorm, FC-21, ReLU, BNorm, FC-5, Softmax}\\
\hline
\multirow{1}{*}{\textbf{Tox21}} & \texttt{FC-4899, ReLU, BNorm, FC-4899, ReLU, BNorm, FC-2, Softmax}\\
\hline
\multirow{1}{*}{\textbf{Diabetes}} & 
\texttt{FC-820, ReLU, BNorm, FC-820, ReLU, BNorm, FC-2, Softmax} \\
\hline
\multirow{2}{*}{\textbf{Hypertension}} & 
\texttt{FC-470, ReLU, BNorm, FC-470, ReLU, BNorm,}\\
& \texttt{FC-470, ReLU, BNorm, FC-2, Softmax} \\
\hline
\multirow{2}{*}{\textbf{Cholesterol}} & 
\texttt{FC-480, ReLU, BNorm, FC-480, ReLU, BNorm, FC-480, ReLU, BNorm,}\\
& \texttt{FC-480, ReLU, BNorm, FC-2, Softmax} \\
\hline
\multirow{2}{*}{\textbf{Landsat}} & 
\texttt{FC-68, ReLU, BNorm, FC-68, ReLU, BNorm, FC-68, ReLU, BNorm,}\\
& \texttt{FC-68, ReLU, BNorm, FC-6, Softmax} \\
\hline
\multirow{2}{*}{\textbf{MIT-BIH}} & 
\texttt{FC-1737, ReLU, BNorm, FC-1737, ReLU, BNorm,}\\
& \texttt{FC-1737, ReLU, BNorm, FC-5, Softmax} \\
\bottomrule
\end{tabular}
}
\end{center}
\end{minipage}
\label{tab:archs_mlp}
\end{table*}%

\begin{table*}[h]%
\renewcommand{\arraystretch}{1.4}
\caption{SNN architectures used in our experiments.}
\begin{minipage}{\textwidth}
\begin{center}
\resizebox{\columnwidth}{!}{
\begin{tabular}{l|l}
\toprule
\textbf{Dataset} & \textbf{Architecture}\\
\hline
\multirow{2}{*}{\textbf{CIFAR-10}} & 
\texttt{FC-3901, SeLU, BNorm, FC-3901, SeLU, BNorm, }\\
& \texttt{FC-3901, SeLU, BNorm, FC-10, Softmax} \\
\hline
\multirow{2}{*}{\textbf{HAPT}} & \texttt{FC-510, ReLU, BNorm, FC-510, SeLU, BNorm, }\\
& \texttt{FC-510, SeLU, FC-5, Softmax} \\
\hline
\multirow{2}{*}{\textbf{Tox21}} & \texttt{FC-3666, ReLU, BNorm, FC-3666, SeLU, BNorm, }\\
& \texttt{FC-3666, SeLU, FC-2, Softmax} \\
\hline
\multirow{1}{*}{\textbf{Diabetes}} & 
\texttt{FC-160, SeLU, BNorm, FC-160, SeLU, BNorm, FC-2, Softmax} \\
\hline
\multirow{1}{*}{\textbf{Hypertension}} & 
\texttt{FC-213, SeLU, BNorm, FC-213, SeLU, BNorm, FC-2, Softmax} \\
\hline
\multirow{2}{*}{\textbf{Cholesterol}} & \texttt{FC-122, ReLU, BNorm, FC-122, SeLU, BNorm, }\\
& \texttt{FC-122, SeLU, FC-2, Softmax} \\
\hline
\multirow{1}{*}{\textbf{Landsat}} & 
\texttt{FC-816, SeLU, BNorm, FC-816, SeLU, BNorm, FC-6, Softmax} \\
\hline
\multirow{2}{*}{\textbf{MIT-BIH}} & 
\texttt{FC-1140, SeLU, BNorm, FC-1140, SeLU, BNorm,}\\
& \texttt{FC-1140, SeLU, BNorm, FC-5, Softmax} \\
\bottomrule
\end{tabular}
}
\end{center}
\end{minipage}
\label{tab:archs_snn}
\end{table*}%

\begin{table*}[h]%
\renewcommand{\arraystretch}{1.4}
\caption{SET architectures used in our experiments.}
\begin{minipage}{\textwidth}
\begin{center}
\resizebox{\columnwidth}{!}{
\begin{tabular}{l|l}
\toprule
\textbf{Dataset} & \textbf{Architecture}\\
\hline
\multirow{1}{*}{\textbf{CIFAR-10}} & 
\texttt{SC-4000, SReLU, SC-1000, SReLU, SC-4000, SReLU, FC-10, Softmax} \\
\hline
\multirow{1}{*}{\textbf{HAPT}} & 
\texttt{SC-500, SReLU, SC-500, SReLU, SC-500, SReLU, FC-5, Softmax} \\
\hline
\multirow{1}{*}{\textbf{Tox21}} & 
\texttt{SC-1000, SReLU, SC-1000, SReLU, SC-1000, SReLU, FC-2, Softmax} \\
\hline
\multirow{1}{*}{\textbf{Diabetes}} & 
\texttt{SC-1000, SReLU, SC-1000, SReLU, SC-1000, SReLU, FC-2, Softmax} \\
\hline
\multirow{1}{*}{\textbf{Hypertension}} & 
\texttt{SC-1000, SReLU, SC-1000, SReLU, SC-1000, SReLU, FC-2, Softmax} \\
\hline
\multirow{1}{*}{\textbf{Cholesterol}} & 
\texttt{SC-1000, SReLU, SC-1000, SReLU, SC-1000, SReLU, FC-2, Softmax} \\
\hline
\multirow{1}{*}{\textbf{Landsat}} & 
\texttt{SC-1000, SReLU, SC-1000, SReLU, SC-1000, SReLU, FC-6, Softmax} \\
\hline
\multirow{1}{*}{\textbf{MIT-BIH}} & 
\texttt{SC-1000, SReLU, SC-1000, SReLU, SC-1000, SReLU, FC-5, Softmax} \\
\bottomrule
\end{tabular}
}
\end{center}
\end{minipage}
\label{tab:archs_set}
\end{table*}%

\begin{table*}[h]%
\renewcommand{\arraystretch}{1.4}
\caption{FGR architectures used in our experiments.}
\begin{minipage}{\textwidth}
\begin{center}
\resizebox{0.8\columnwidth}{!}{
\begin{tabular}{l|l}
\toprule
\textbf{Dataset} & \textbf{Architecture}\\
\hline
\multirow{1}{*}{\textbf{CIFAR-10}} & 
\texttt{Group-256, FC-3072, ReLU, FC-10, Softmax} \\
\hline
\multirow{1}{*}{\textbf{HAPT}} & 
\texttt{Group-104, FC-12173, ReLU, FC-5, Softmax}\\
\hline
\multirow{1}{*}{\textbf{Tox21}} & 
\texttt{Group-3468, FC-3468, ReLU, FC-2, Softmax}\\
\hline
\multirow{1}{*}{\textbf{Diabetes}} & 
\texttt{Group-100, FC-230, ReLU, FC-2, Softmax}\\
\hline
\multirow{1}{*}{\textbf{Hypertension}} & 
\texttt{Group-100, FC-210, ReLU, FC-2, Softmax}\\
\hline
\multirow{1}{*}{\textbf{Cholesterol}} & 
\texttt{Group-100, FC-250, ReLU, FC-2, Softmax}\\
\hline
\multirow{1}{*}{\textbf{Landsat}} & 
\texttt{Group-32, FC-1577, ReLU, FC-6, Softmax}\\
\hline
\multirow{1}{*}{\textbf{MIT-BIH}} & 
\texttt{Group-160, FC-3444, ReLU, FC-5, Softmax}\\

\bottomrule
\end{tabular}
}
\end{center}
\end{minipage}
\label{tab:archs_fgr}
\end{table*}%

\clearpage
\section{Software Implementation}

Table~\ref{tab:swdeps} presents the list of software dependencies and versions used in our implementation. To produce results related to this paper, we used a workstation with 4 NVIDIA GeForce RTX-2080Ti GPUs, a 12 core Intel \mbox{Core i9-7920X} processor, and 128 GB memory. Each experiment took between about 30 minutes to 72 hours, based on the task and method being tested.

\begin{table}[h]%
\renewcommand{\arraystretch}{1.3}
\caption{Software dependencies.}
\begin{center}
\texttt{
\begin{tabular}{l|l}
\toprule
\textbf{Dependency} & \textbf{Version} \\
\hline
python & 3.7.1 \\
pytorch & 1.1.0 \\
torchvision & 0.2.1 \\
cuda100 & 1.0 \\
ipython & 6.5.0 \\
jupyter & 1.0.0 \\
numpy & 1.15.4 \\
nni & 0.9.1.1 \\
pandas & 0.23.4 \\
scikit-learn & 0.19.2 \\
scipy & 1.1.0 \\
pomegranate & 0.11.1 \\
tqdm & 4.32.1 \\
matplotlib & 3.0.1 \\
\bottomrule
\end{tabular}}
\end{center}
\label{tab:swdeps}
\end{table}%

\clearpage
\section{Experiments on MNIST and Synthesized Data}
\label{sec:mnist-synthesized}
\begin{figure}[h]
    \centering
    \begin{subfigure}[b]{0.21\columnwidth}
        \centering
        \includegraphics[width=\columnwidth]{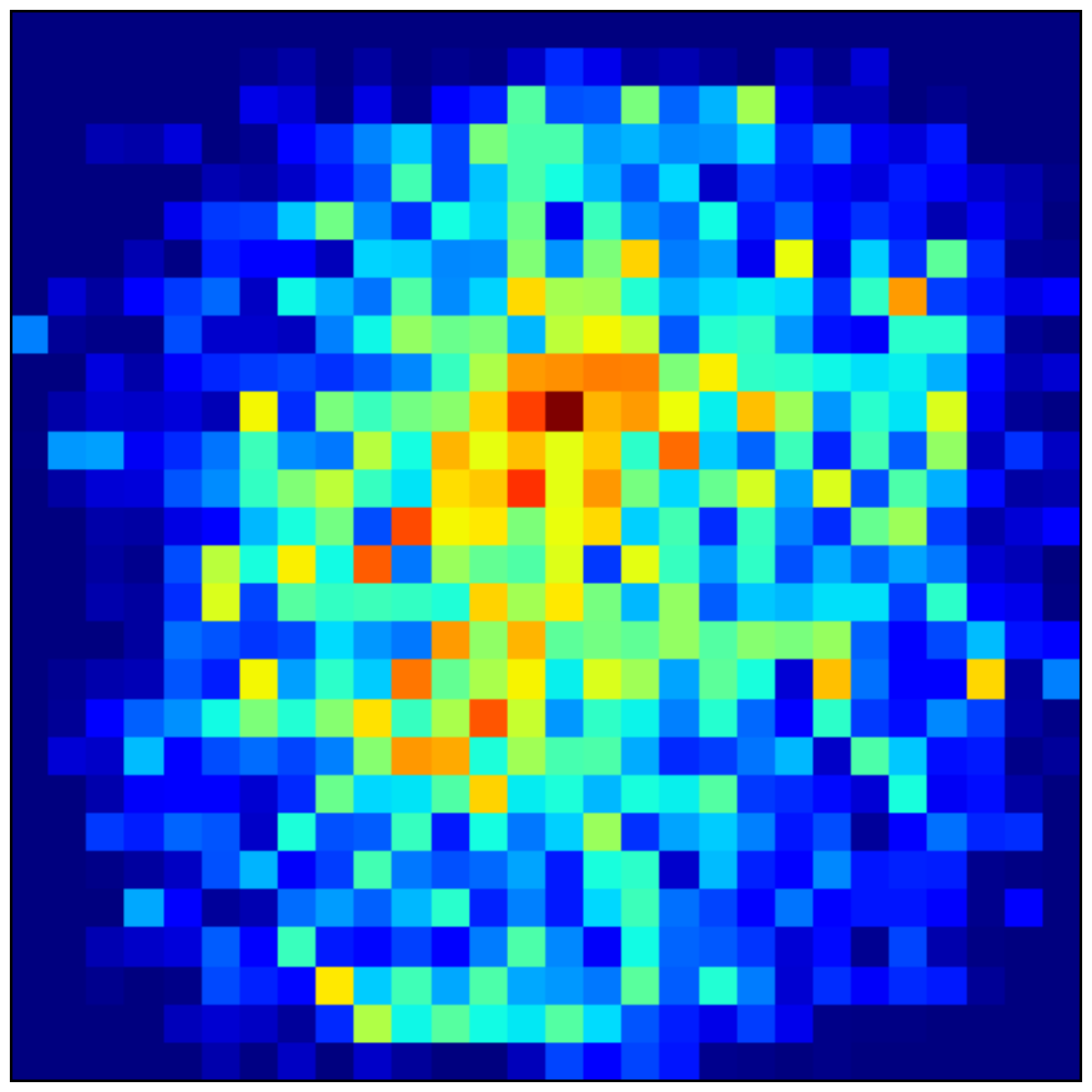}
        \caption{}
        \label{fig:vis_sum_mnist_full}
    \end{subfigure}
    ~
    \begin{subfigure}[b]{0.21\columnwidth}
        \centering
        \includegraphics[width=\columnwidth]{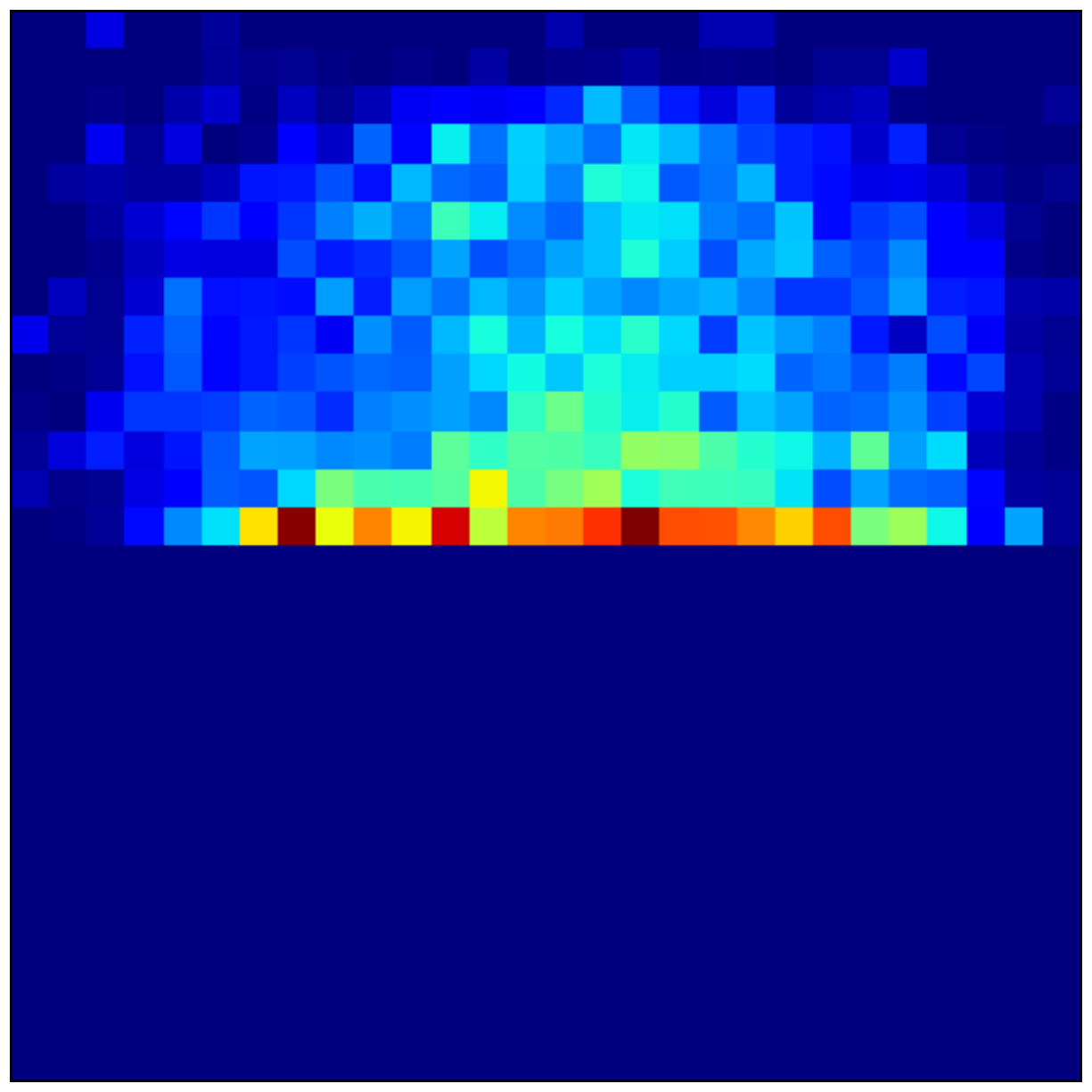}
        \caption{}
        \label{fig:vis_sum_mnist_half}
    \end{subfigure}
    \caption{MNIST visualization of pixels selected by the \texttt{Group-Select} layer: (a) using complete images as input, (b) using images that the lower half is replaced by Gaussian noise. In this figure, warmer colors represent pixels being being present in more groups.}
    \label{fig:vis_sum_mnist}
    \vspace{-0.0in}
\end{figure}

MNIST dataset is used to visually inspect the performance of the \texttt{Group-Select} layer. Figure~\ref{fig:vis_sum_mnist} shows a heat-map of how frequently each pixel is selected across all feature groups for: $(a)$ original MNIST samples, $(b)$ MNIST samples where the lower-half is replaced by Gaussian noise. From Figure~\ref{fig:vis_sum_mnist_full}, it can be seen that most groups are selecting pixels within the center of the frame, effectively discarding margin pixels. This is consistent with other work which show the importance of different locations for MNIST images\footnote{Kachuee, M., Darabi, S., Moatamed, B., \& Sarrafzadeh, M. (2018). Dynamic feature acquisition using denoising autoencoders. IEEE transactions on neural networks and learning systems, 30(8), 2252-2262.}. Apart from this, in Figure~\ref{fig:vis_sum_mnist_half}, a version of the MNIST dataset is used in which half of the frame does not provide any useful information for the downstream classification task. From this figure, GMLP is not selecting any features to be used from the lower region.

\begin{figure}[h]
\centering
\begin{minipage}[t]{0.35\columnwidth}
    \centering
        \includegraphics[trim=0 30 0 0,clip,width=\columnwidth]{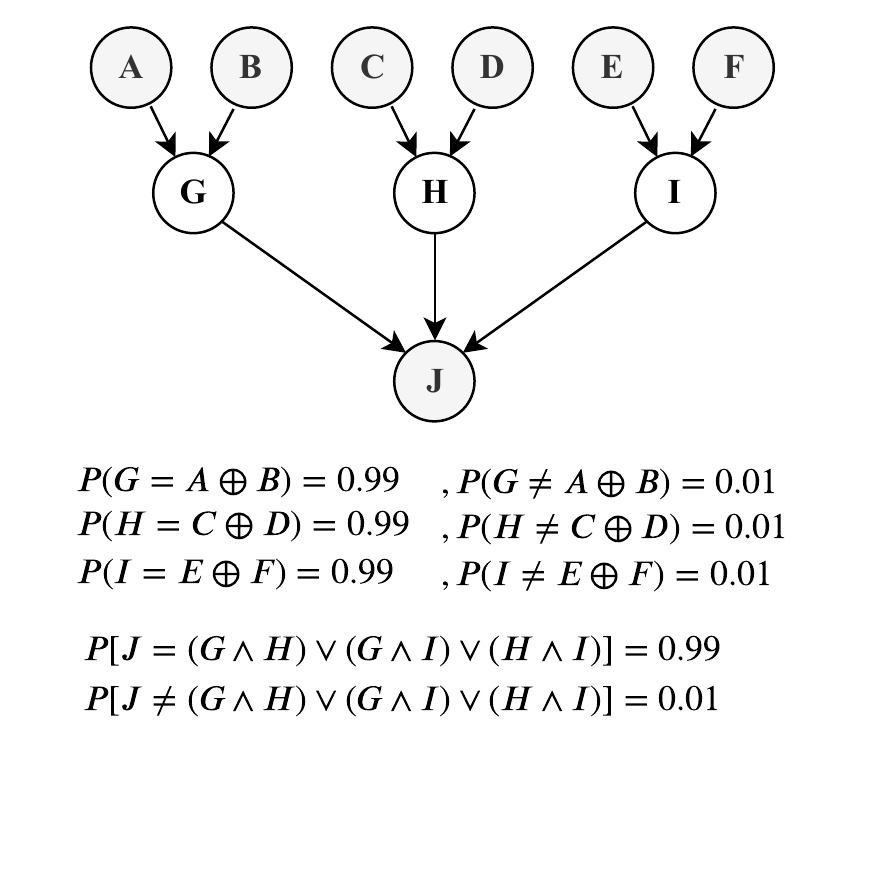}
        \caption{The Bayesian network and conditionals used to generate the synthesized dataset of binary features A-F and target J.}
        \label{fig:synthesizedbn}
\end{minipage}
~
\begin{minipage}[t]{0.3\columnwidth}
    \centering
        \includegraphics[width=\columnwidth]{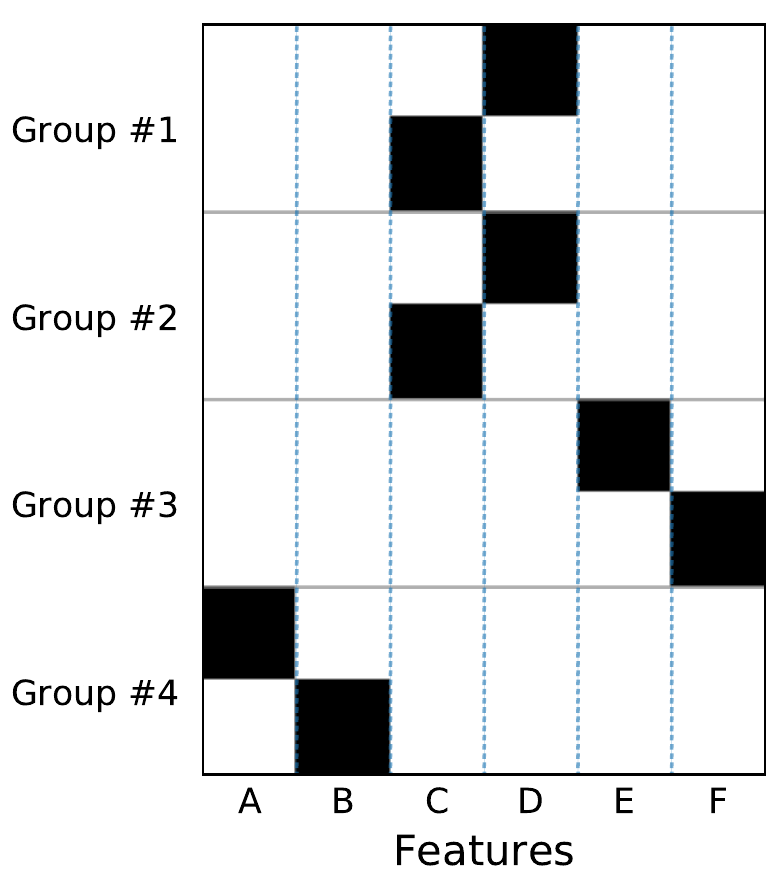}
        \caption{Visualization of the selected features within each group. Every two consecutive rows show features selected for a certain group.}
        \label{fig:vis_groups_synthesizedbn}
\end{minipage}
\end{figure}

In order to show the effectiveness of GMLP, we synthesized a dataset which has intrinsic and known expressive feature groups. Specifically, we used a simple Bayesian network as depicted in Figure~\ref{fig:synthesizedbn}. This network consists of six binary features, A to F, interacting with each other as specified by the graph edges, which determine the distribution of the target node, J. The graph and conditionals are designed such that each of the nodes in the second level take the XOR value of their parents with a $99\%$ probability. The target node, J, is essentially one with a high probability if at least two of the second level nodes are one. We synthesized dataset by sampling 6,400 samples from the network (1,280 samples for test and the rest of training/evaluation). On this dataset, we trained a very simple GMLP consisting of four groups of size two, one group-wise fully-connected layer, and an output layer. Figure~\ref{fig:vis_groups_synthesizedbn} shows the features selected for each group after the training phase (i.e., the $\Psi$ matrix). From this figure, the \texttt{Group-Select} layer successfully learns to detect the feature pairs that are interacting, enabling the \texttt{Group-FC} layers to decode the non-linear XOR relations.

\begin{figure}[h]
        \centering
        \includegraphics[width=0.7\columnwidth]{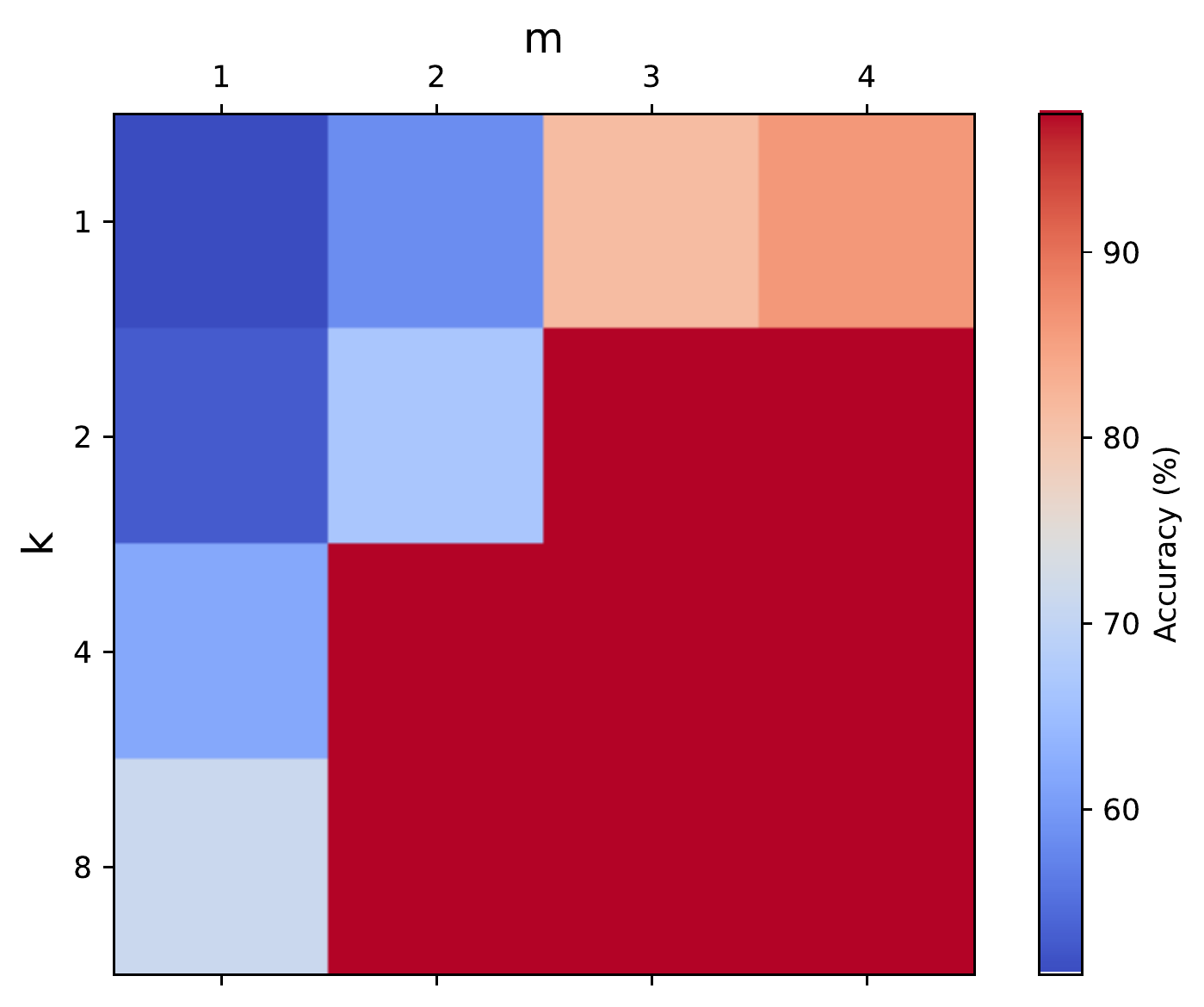}
        \caption{Visualization of prediction accuracies for the synthesized dataset using different group size (m) and number of groups (k).}
        \label{fig:acc_synthesizedbn}
\end{figure}

To investigate the impact of GMLP architectures that does match the data generating distribution, we conducted experiments by changing the group size and number of groups. Based on the results presented in Figure~\ref{fig:acc_synthesizedbn}, any network with $m$ and $k$ values more than 2 is able to fit the distribution. However, in this example, using smaller $m$ and $k$ values results in a significant degradation due to the incapability of the networks to capture the dataset dynamics. This result demonstrates the importance of the $m$ and $k$ hyperparameters.

\clearpage
\section{Additional Ablation Studies}
\begin{figure}[h]
\centering
\begin{minipage}[t]{0.45\columnwidth}
    \centering
        \includegraphics[width=\columnwidth]{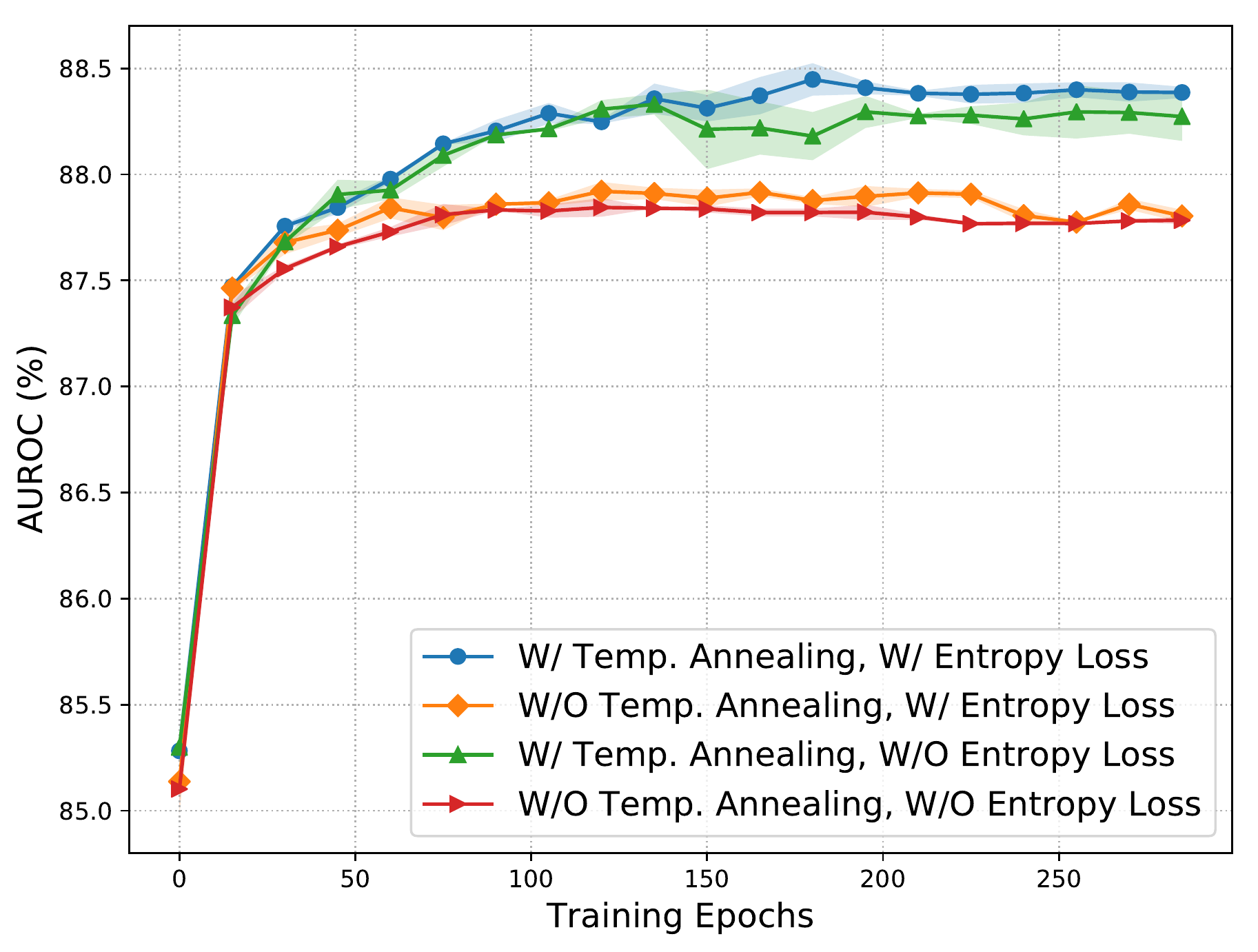}
        \caption{Ablation study on the impact of temperature annealing and entropy loss terms.}
        \label{fig:ablation_diabetes_temphloss}
\end{minipage}
~
\begin{minipage}[t]{0.45\columnwidth}
    \centering
        \includegraphics[width=\columnwidth]{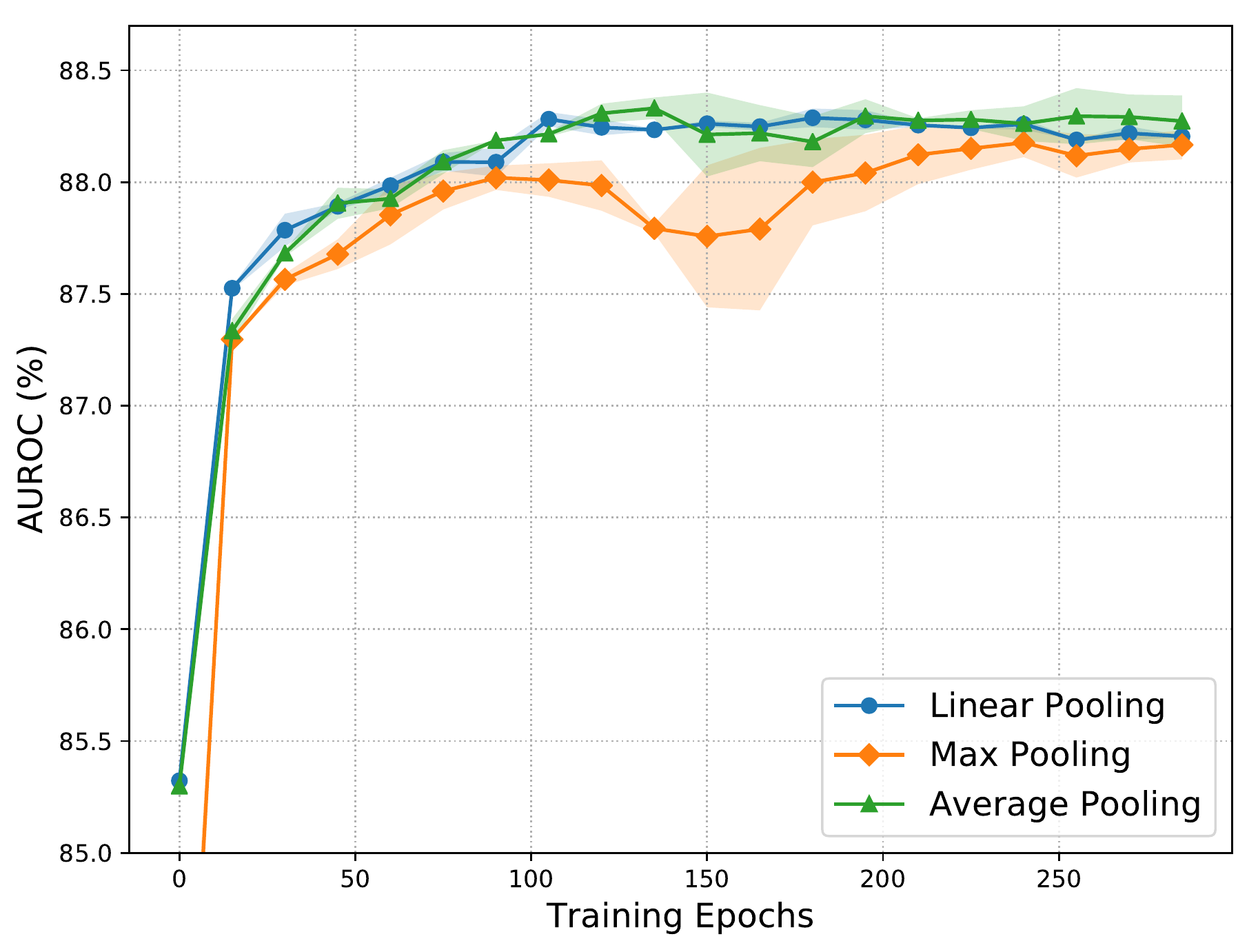}
        \caption{Ablation study demonstrating the impact of different pooling functions.}
        \label{fig:ablation_diabetes_pooling}
\end{minipage}
\vspace{-0.1in}
\end{figure}

Figure~\ref{fig:ablation_diabetes_temphloss} presents an ablation study comparing the performance of GMLP on the Diabetes dataset for networks trained: $(i)$  using both the temperature annealing and the entropy loss objective, $(ii)$ using only temperature annealing without the entropy loss objective, $(iii)$ using no temperature annealing but using the entropy loss objective, $(iv)$ not using any of the temperature annealing or the entropy loss objective.
From this figure, excluding both techniques leads to a significantly lower performance. Also note that, compared to the CIFAR-10 ablation experiments (see Figure~\ref{fig:ablation_cifar10_temphloss}), only using the entropy loss term is not sufficient for achieving best results. We found that using both techniques consistently achieves better or similar results.

Figure~\ref{fig:ablation_diabetes_pooling} shows a comparison between GMLP models trained on the Diabetes dataset using different pooling types: $(i)$ linear transformation, $(ii)$ max pooling, and $(iii)$ average pooling. As it can be seen from this comparison, while there are slight differences in the convergence speed of using different pooling types, all of them achieve relatively similar accuracies.

\begin{figure}[h]
\centering
\begin{minipage}[t]{0.42\columnwidth}
    \centering
        \includegraphics[width=\columnwidth]{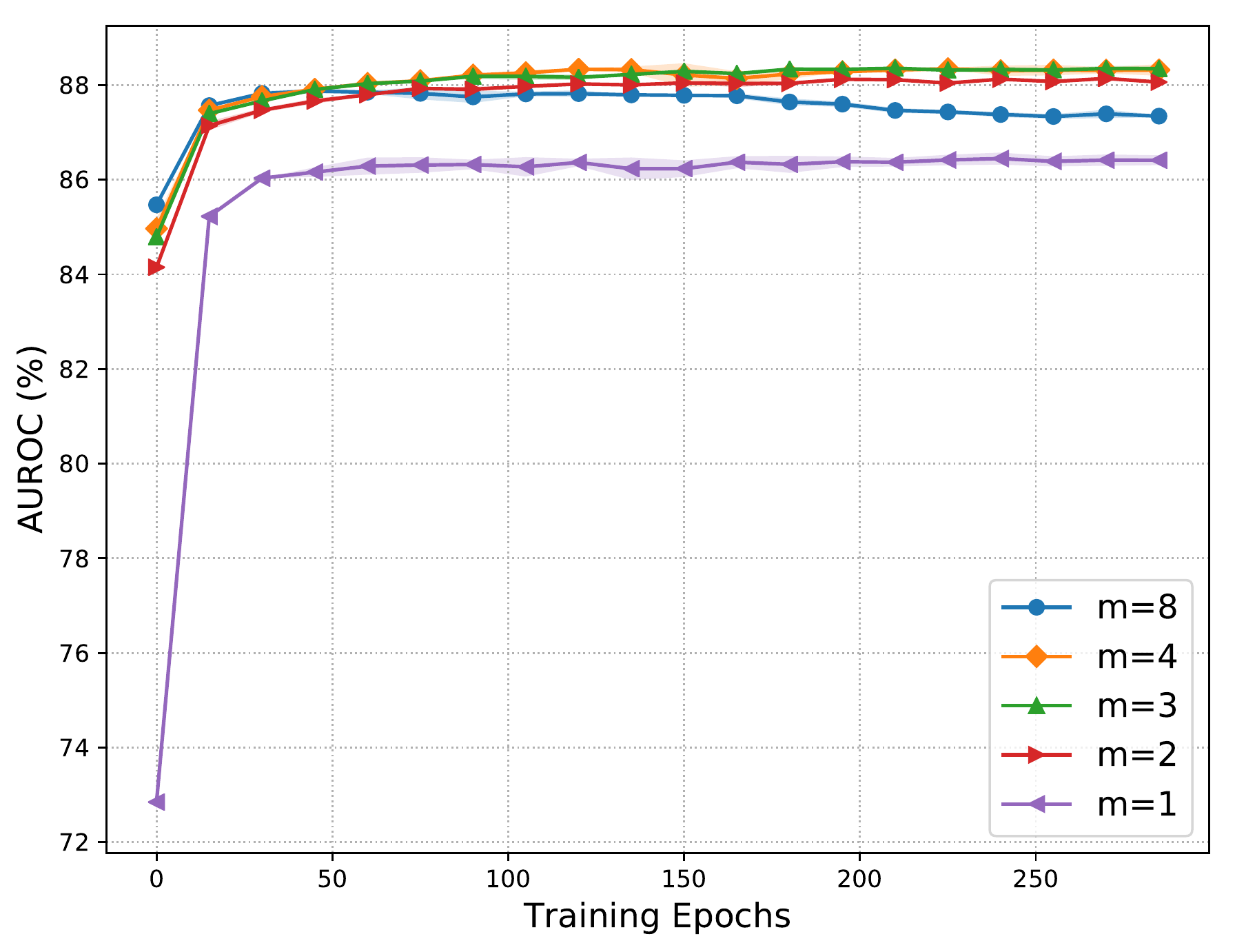}
        \caption{Ablation study on the impact of using different group sizes ($m$). For this experiment, we used $k$=352. }
        \label{fig:ablation_diabetes_m}
\end{minipage}
~
\begin{minipage}[t]{0.42\columnwidth}
    \centering
        \includegraphics[width=\columnwidth]{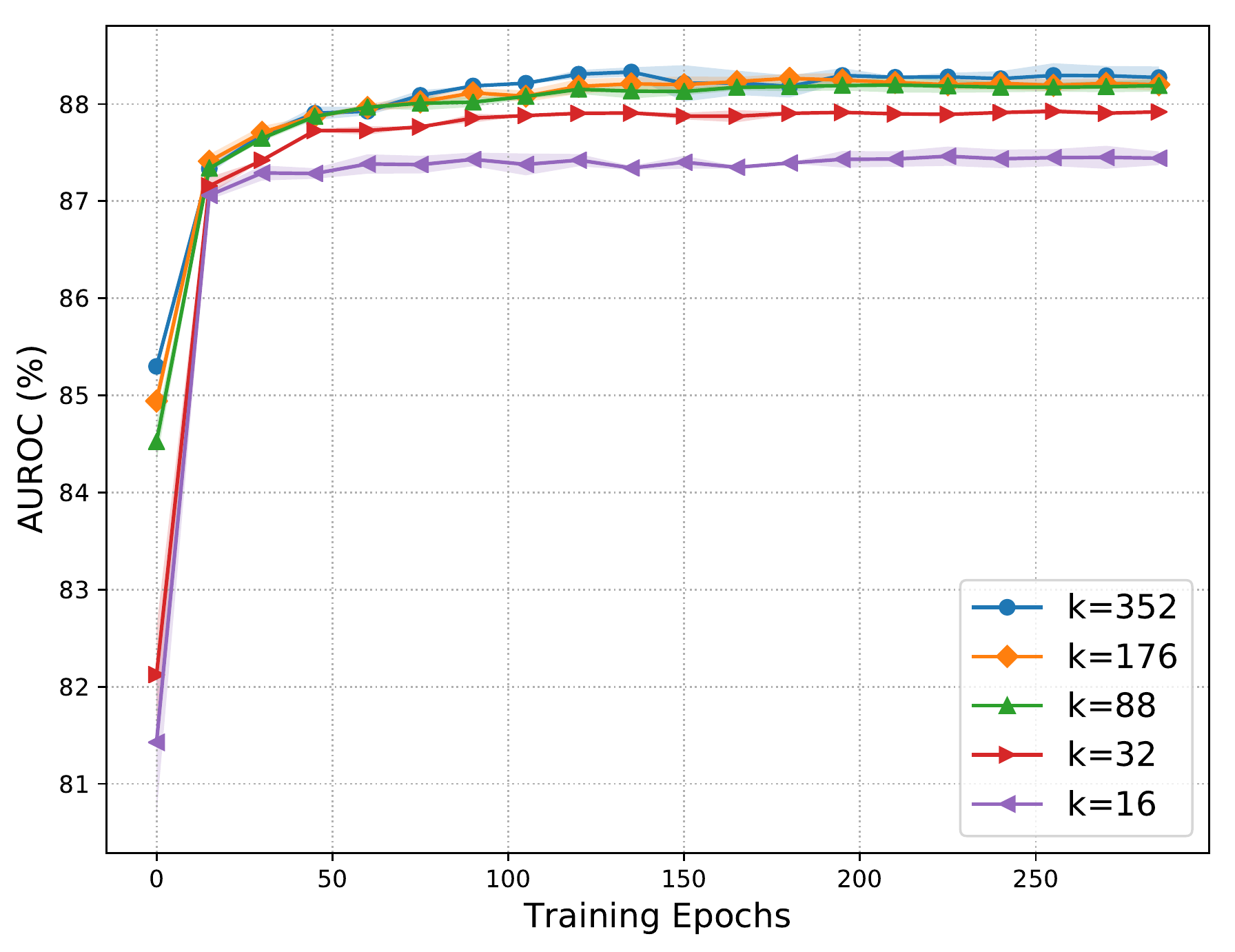}
        \caption{Ablation study on the impact of using different number of groups ($k$). For this experiment, we used $m$=4.}
        \label{fig:ablation_diabetes_k}
\end{minipage}
\vspace{-0.1in}
\end{figure}

Figure~\ref{fig:ablation_diabetes_m} shows learning curves for training the Diabetes models using different group sizes. From this figure, using very small group sizes decreases the final accuracy. At the other extreme, the improvement achieved using larger values is negligible for $m$ values more than 3. On the other hand, using very large values degrades the results due to overfitting. Figure~\ref{fig:ablation_diabetes_k} shows a comparison between learning curves for using a different number of groups. Using very small $k$ values result in a significant reduction in performance. However, the rate of performance gains for using more groups is very small for $k$ of more than 176.

\clearpage
\section{Experiments using Alternative Tree Structures}
\begin{figure}[h]
\centering
\begin{minipage}[t]{0.42\columnwidth}
    \centering
        \includegraphics[width=\columnwidth]{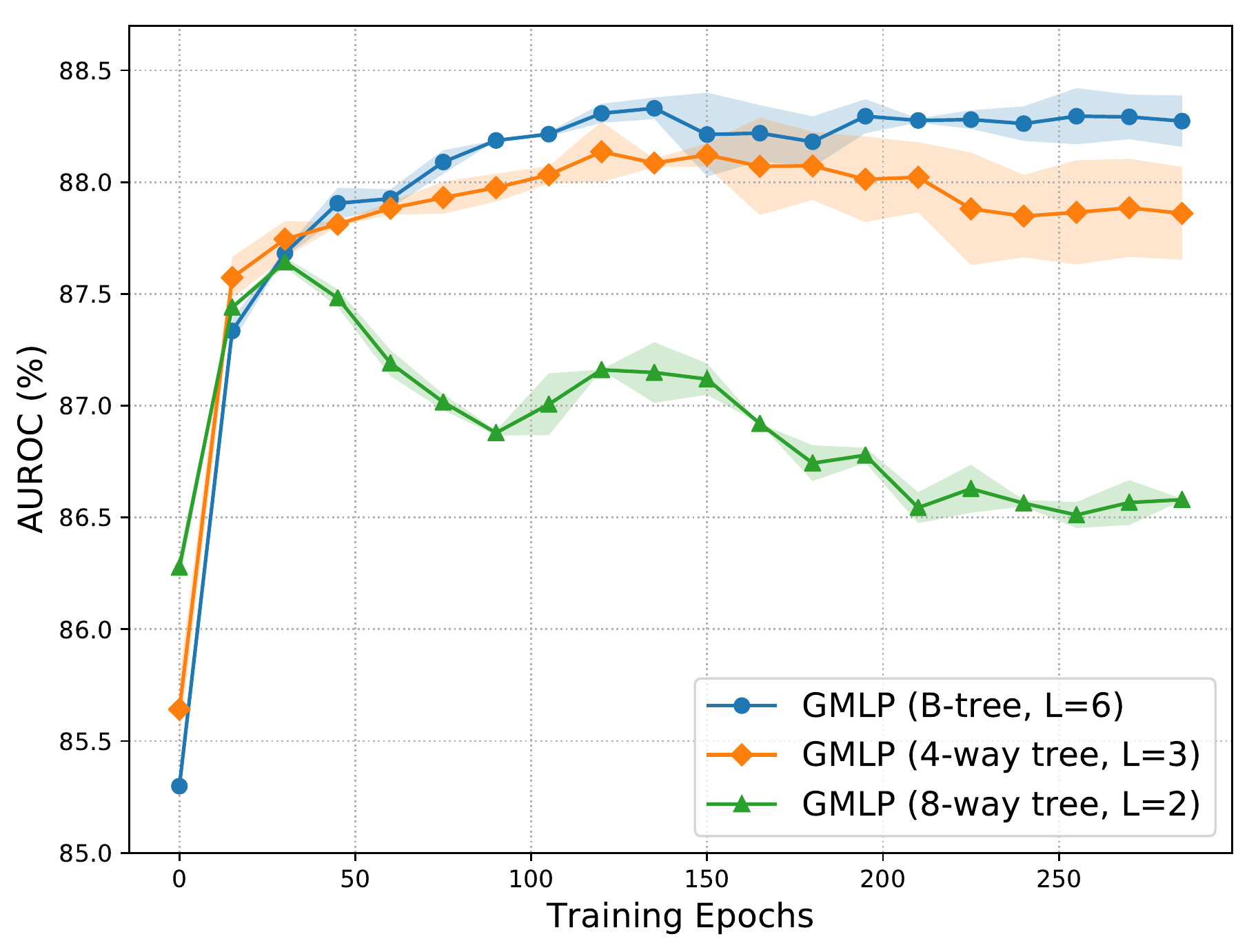}
        \caption{The impact of using different tree types. For this experiment, we used $k$=352 and adjusted the number of layers.}
        \label{fig:ablation_diabetes_arch_K352}
\end{minipage}
~
\begin{minipage}[t]{0.42\columnwidth}
    \centering
        \includegraphics[width=\columnwidth]{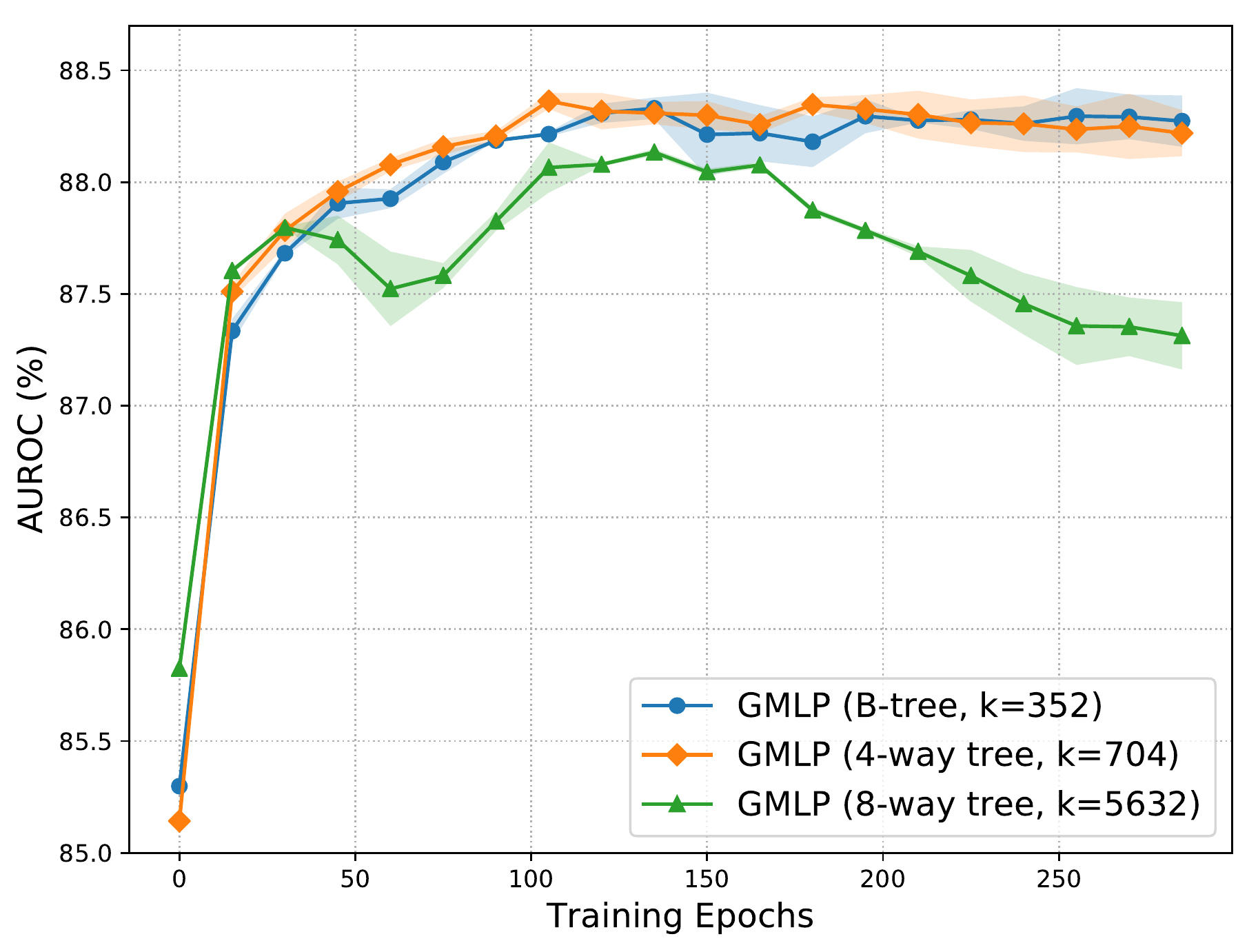}
        \caption{The impact of using different tree types. For this experiment, we used $L$=6 and adjusted the number of groups.}
        \label{fig:ablation_diabetes_arch_L6}
\end{minipage}
\end{figure}

We conducted experiments comparing the suggested binary tree (B-tree) GMLP architecture with other alternatives using tree structures with different branching factors (tree ways). As the GMLP baseline for the Diabetes dataset has 352 groups and 6 layers, increasing the branching factor necessitates either increasing the number of groups or reducing the number of layers to build the tree. To investigate this, we conducted to experiments. First, in Figure~\ref{fig:ablation_diabetes_arch_K352} we fixed the number of groups and experimented on b-tree, 4-way tree, and 8-way tree architectures. Second, in Figure~\ref{fig:ablation_diabetes_arch_L6} we fixed the number of layers and experimented on b-tree, 4-way tree, and 8-way tree architectures. From these results, the B-tree architecture appears to consistently show better or similar results, supporting the use of B-tree architectures for GMLP experiments in this paper. Also, from Figure~\ref{fig:ablation_diabetes_arch_K352}~and~\ref{fig:ablation_diabetes_arch_L6}, using aggressively large branching factors results in a significant performance degradation as pooling many groups together results in information loss.

We hypothesize that the B-tree architecture outperforms other alternatives due to two factors: $(i)$ In a B-tree, each pooling operation only merges information from two groups whereas for larger branching factor where multiple intermediate representations are being combined this pooling operation may result in larger information loss. $(ii)$ Using larger branching factors, results in an exponentially faster merging of the groups; therefore, limiting us to a much shallower network assuming the same number of groups.

\clearpage
\section{Analysis of the Selected Feature Groups}
\begin{figure}[h]
        \centering
        \includegraphics[width=0.9\columnwidth]{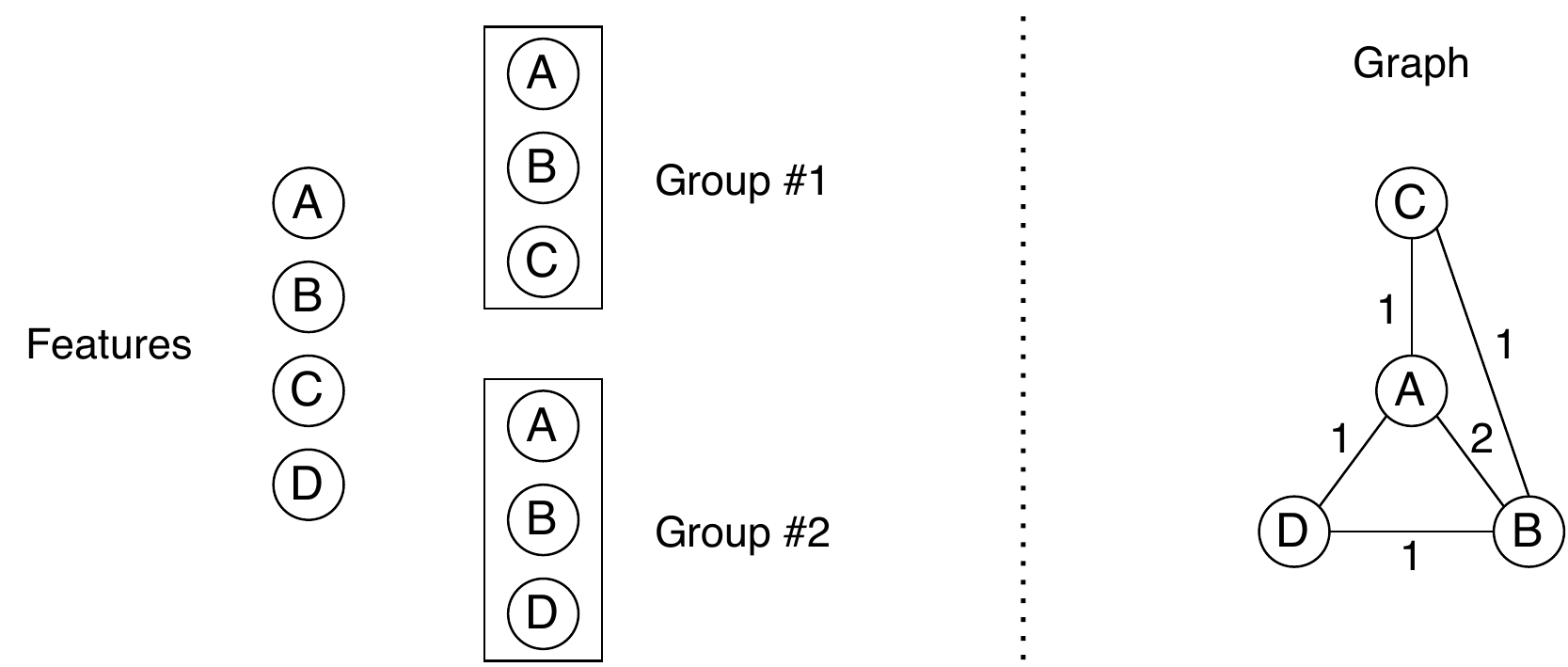}
        \caption{A toy example demonstrating the representation of feature groups as an undirected weighted graph.}
        \label{fig:graph_sample}
\end{figure}

We used undirected graph visualizations to illustrate the feature groups and their relationship in GMLP networks. Specifically, we consider each feature as a graph node and groups as dense connection patterns between the nodes. Here, reappearance of a certain edge, i.e. same sets of features appearing in multiple groups, is considered by an increase in the edge weight. See Figure~\ref{fig:graph_sample} for a toy example demonstrating the representation of feature groups as an undirected weighted graph. To visualize the resulting graphs, we used the spectral graph visualization method from the NetworkX library~\footnote{https://networkx.org} which clusters nodes based on the eigenvectors of the graph Laplacian~\footnote{Von Luxburg, Ulrike. "A tutorial on spectral clustering." Statistics and computing 17.4 (2007): 395-416.}.

\begin{figure}[h]
    \centering
    \begin{subfigure}[b]{0.6\columnwidth}
        \centering
        \includegraphics[width=\columnwidth]{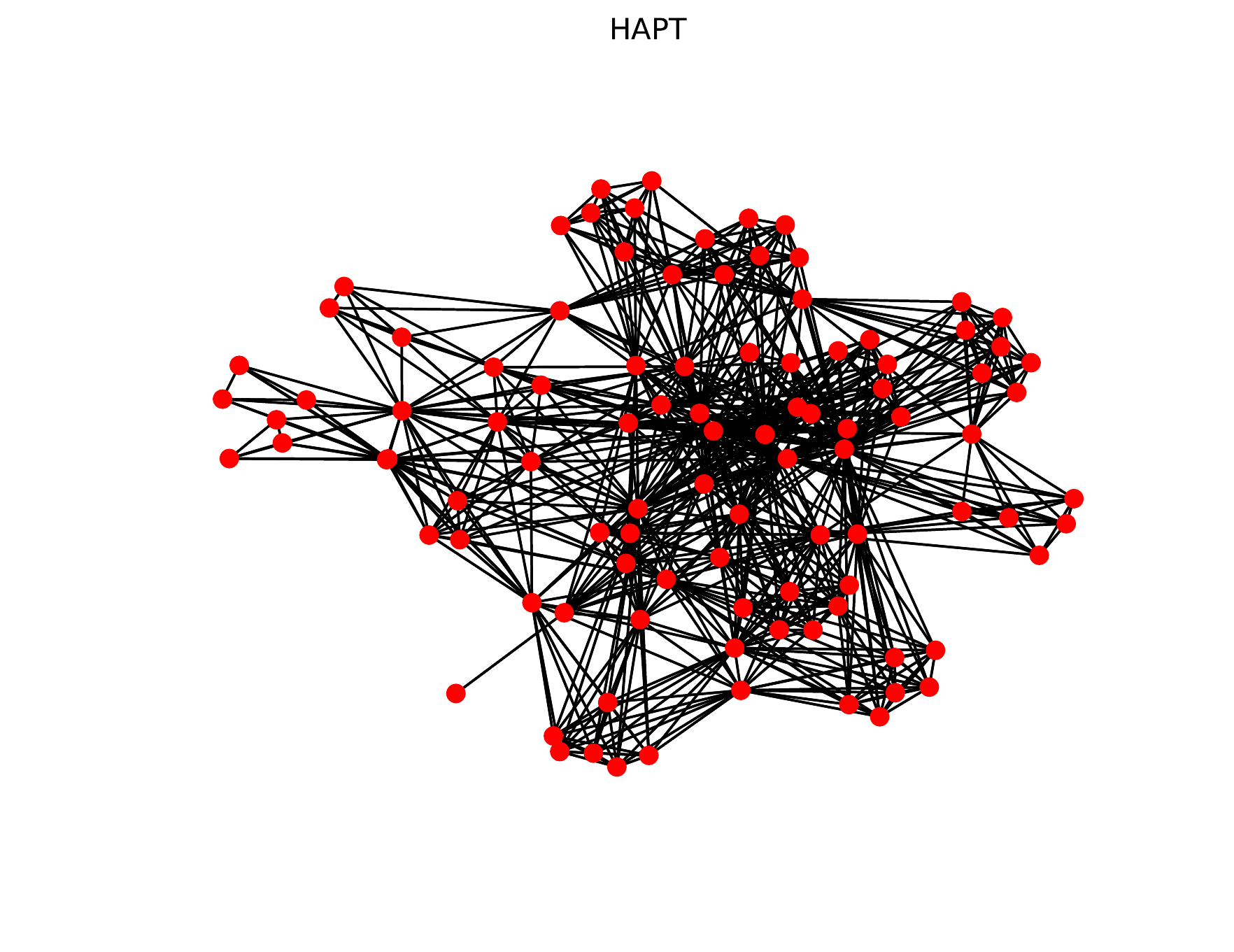}
        \caption{}
        \label{fig:vis_graph_HAPT}
    \end{subfigure}
    
    \begin{subfigure}[b]{0.6\columnwidth}
        \centering
        \includegraphics[width=\columnwidth]{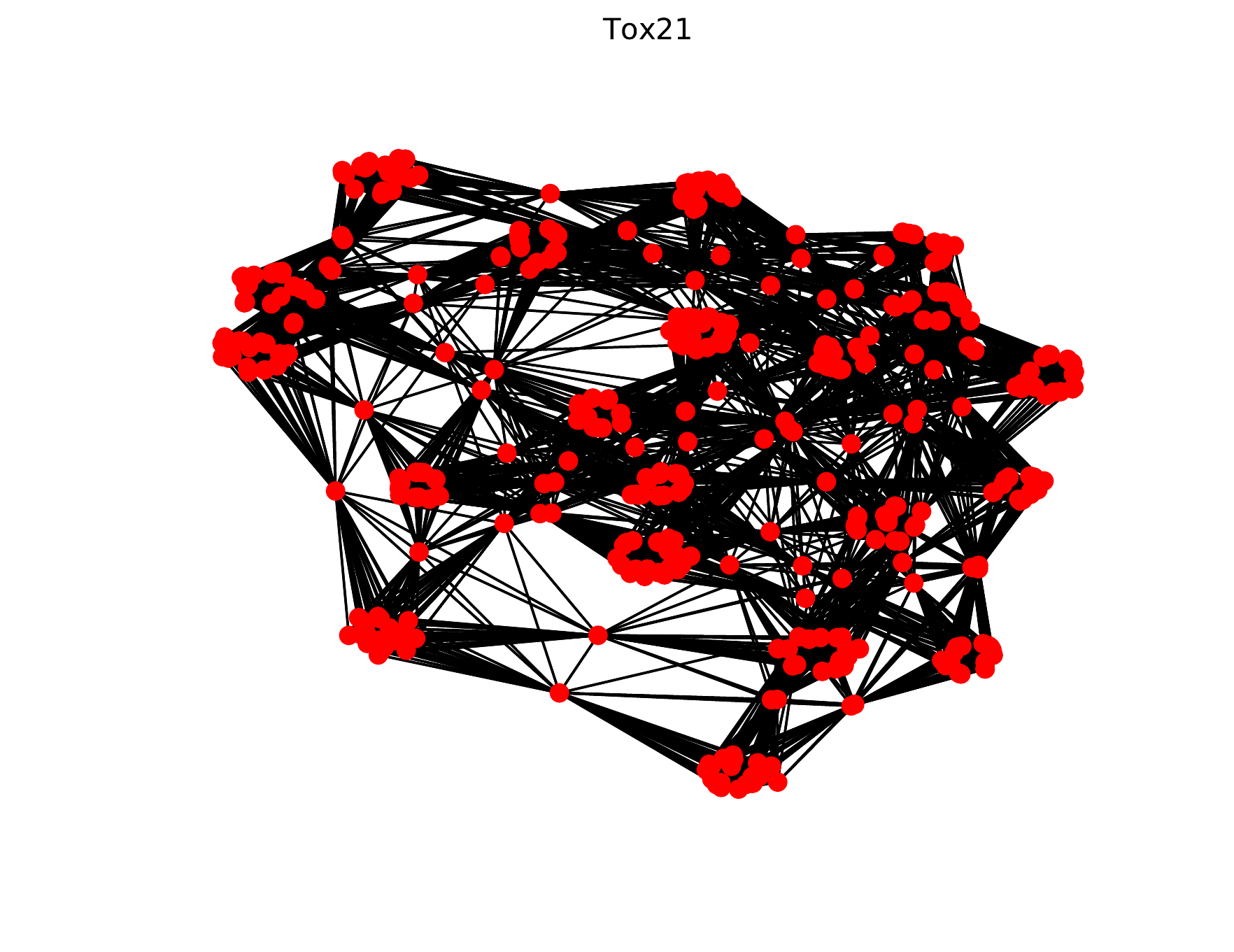}
        \caption{}
        \label{fig:vis_graph_Tox21}
    \end{subfigure}
    
    \begin{subfigure}[b]{0.6\columnwidth}
        \centering
        \includegraphics[width=\columnwidth]{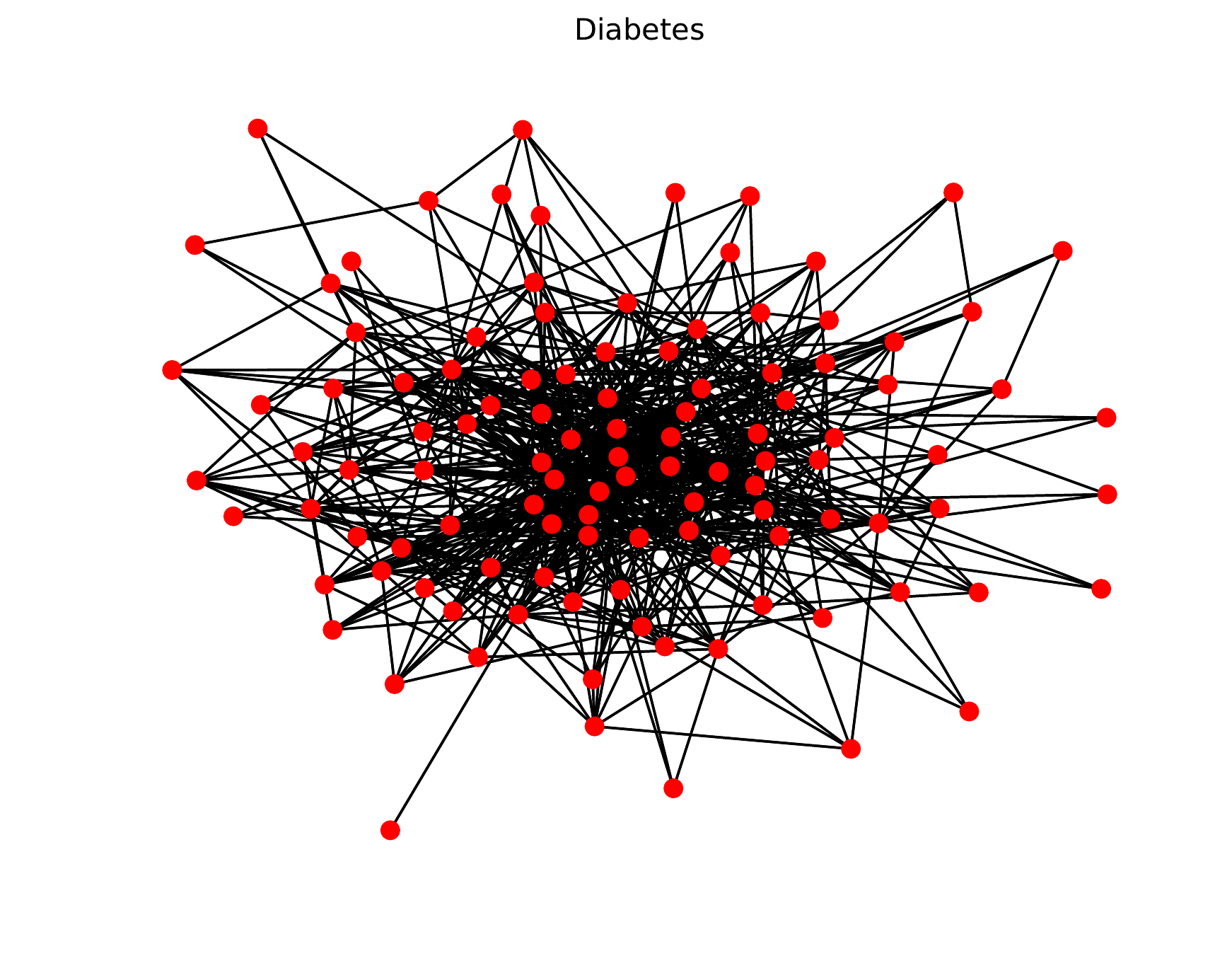}
        \caption{}
        \label{fig:vis_graph_Diabetes}
    \end{subfigure}
    
    \caption{Undirected graphs showing features as nodes and groups as the strength of connection between the nodes for: (a) HAPT, (b) Tox21, and (c) Diabetes datasets.}
    \label{fig:vis_graph}
    \vspace{-0.0in}
\end{figure}

Figure~\ref{fig:vis_graph} presents the resulting graph visualizations for the HAPT, Tox21, and Diabetes datasets. The nodes with the highest weighted connections are clustered and placed in close proximity of each other in the visualization. From this figure, we can observe different grouping patterns for each dataset. For the HAPT dataset, the groups appear to be clustered but have strong overlaps with adjacent clusters. This indicates existence of feature groups that often share features among them. However, for the Tox21 dataset, we observe very strong group clusters with very limited connection to other clusters. Hence, we can conclude that most feature groups are unique and less frequently share features with other groups. The pattern for the Diabetes dataset is very different as certain features are appearing in many groups resulting in a large cluster in the center of the graph, while at the same time there are features that are contributing in very limited number of groups appearing far from the center of the graph. This analysis shows that the feature groups are highly dataset dependent, and using proper hyperparameters, GMLP is able to learn the feature groups.

\clearpage
\section{Visual Analysis}
\label{sec:Visual Analysis}

\begin{wrapfigure}{c}{1.0\linewidth}

\includegraphics[width=\linewidth]{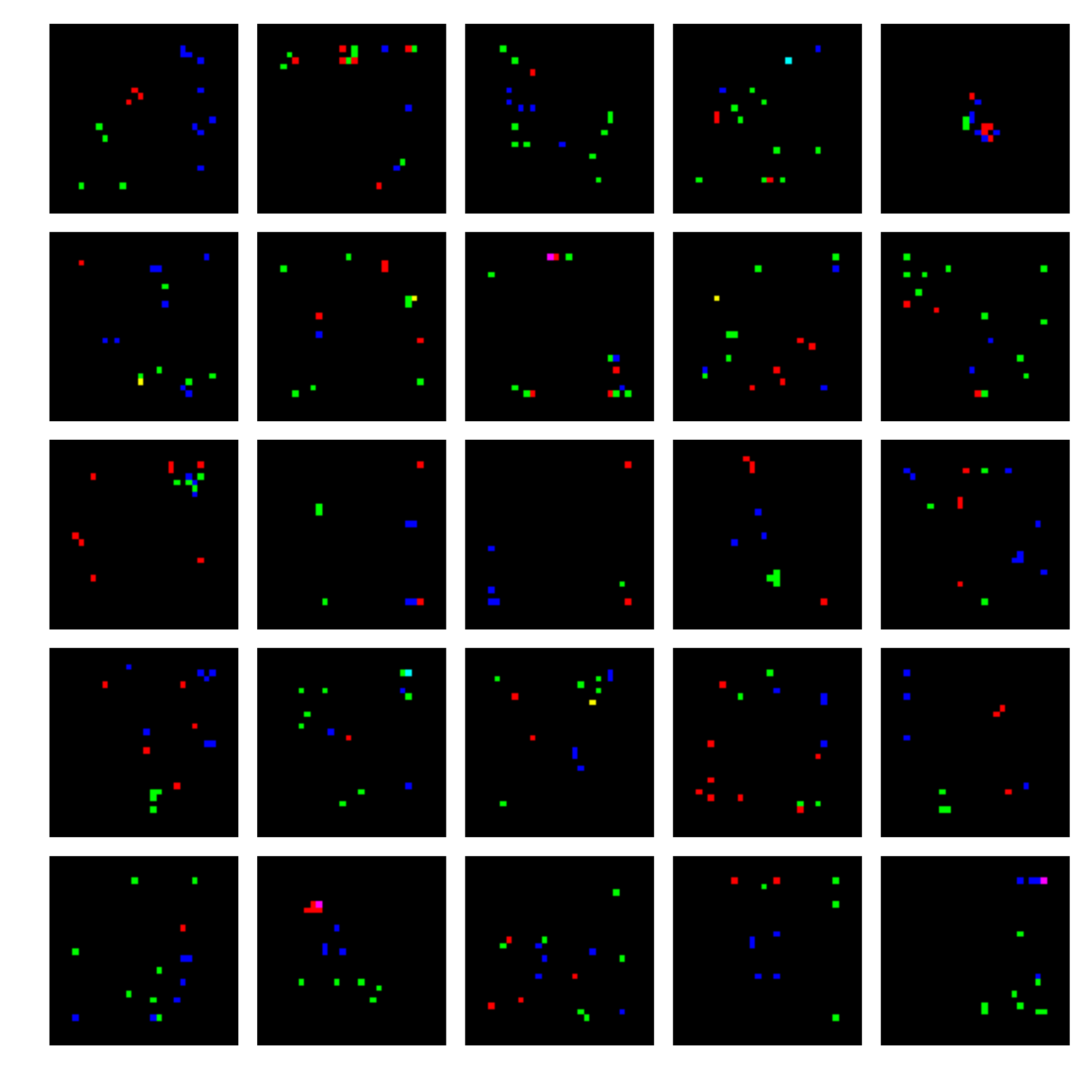}
\caption{Visualization of pixels selected by each group for the CIFAR-10 GMLP architecture. Red, green, and blue colors indicate which channel is selected for each location. Due to space limitations, 25 random groups out of 1536 total groups visualized here.}
        \label{fig:vis_groups_cifar10}
\end{wrapfigure}

In Figure~\ref{fig:vis_groups_cifar10}, we present a visualization of the selected feature for 25 randomly selected groups in our final CIFAR-10 architecture. Red, green, and blue colors indicate which channel is selected for each location. Compared to visualizations that are frequently used for convolutional networks, as GMLP has the flexibility to select pixels at different locations and different color channels, it is not easy to find explicit patterns in this visualization. However, one noticeable pattern is that features selected from a certain color channel usually appear in clusters resembling irregularly shaped patches.


\begin{figure}[h]
        \centering
        \includegraphics[width=0.3\columnwidth]{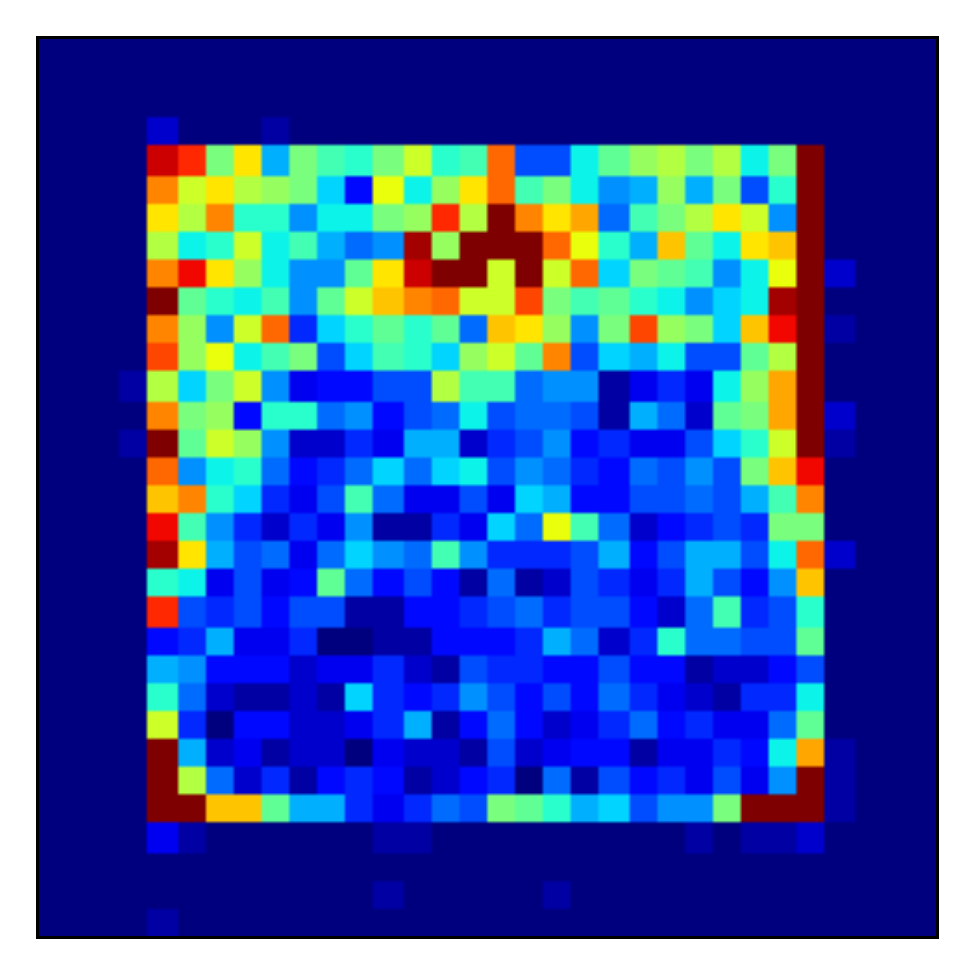}
        \caption{Visualization of pixels selected by the \texttt{group-select} layer for the CIFAR-10 GMLP model. Warmer colors represent features that are being selected more frequently.}
        \label{fig:vis_sum_cifar10}
\end{figure}

\clearpage
Figure~\ref{fig:vis_sum_cifar10} shows the frequency in which each CIFAR-10 location is selected by the GMLP network. From this visualization, GMLP is mostly ignoring the border areas which can be a result of the data augmentation process used to train the network i.e., randomly cropping the center area and padding the margins.

\clearpage
\section{Analysis of the GMLP Training Complexity}
Section~\ref{sec:Analysis} provided a complexity analysis for the GMLP at the prediction time. To extend that analysis to the training time complexity, we need to consider the fact that, at the early stages of the training the routing matrix $\Psi$ is not necessarily sparse. Therefore, the training time memory and compute complexity for a GMLP would be (for simplicity, ignoring bias and pooling terms):
\begin{equation}
    kmd + km^2 + \frac{km^2}{2} + \frac{km^2}{4} + ... + \frac{km^2}{2^{L-1}} + C\frac{km}{2^{L-1}}.
\end{equation}
Therefore, the training complexity is of order $\mathcal{O}(kmd+km^2+\frac{Ckm}{2^{L-1}})$.

In terms of the model size, during the training (before $\Psi$ converges to a sparse matrix), training the GMLP model requires storage for the $km \times d$ routing matrix as well as group-wise fully-connected layers. Based on our experiments, in our implementation, the size of the routing matrix usually plays the dominant role in determining the memory requirements. Nonetheless, we would like to note that we were able to run all experiments in this paper on a mid-range GPU with 11GB memory.

At prediction time, utilizing the sparsity of the routing matrix, the first term reduces to $km$ resulting in a compute and memory complexity of $\mathcal{O}(km^2+\frac{Ckm}{2^{L-1}})$ as suggested in Section~\ref{sec:Analysis}.
\\\\\\\\\\\\
\begin{figure*}[ht]
    \centering
        \includegraphics[width=0.7\linewidth]{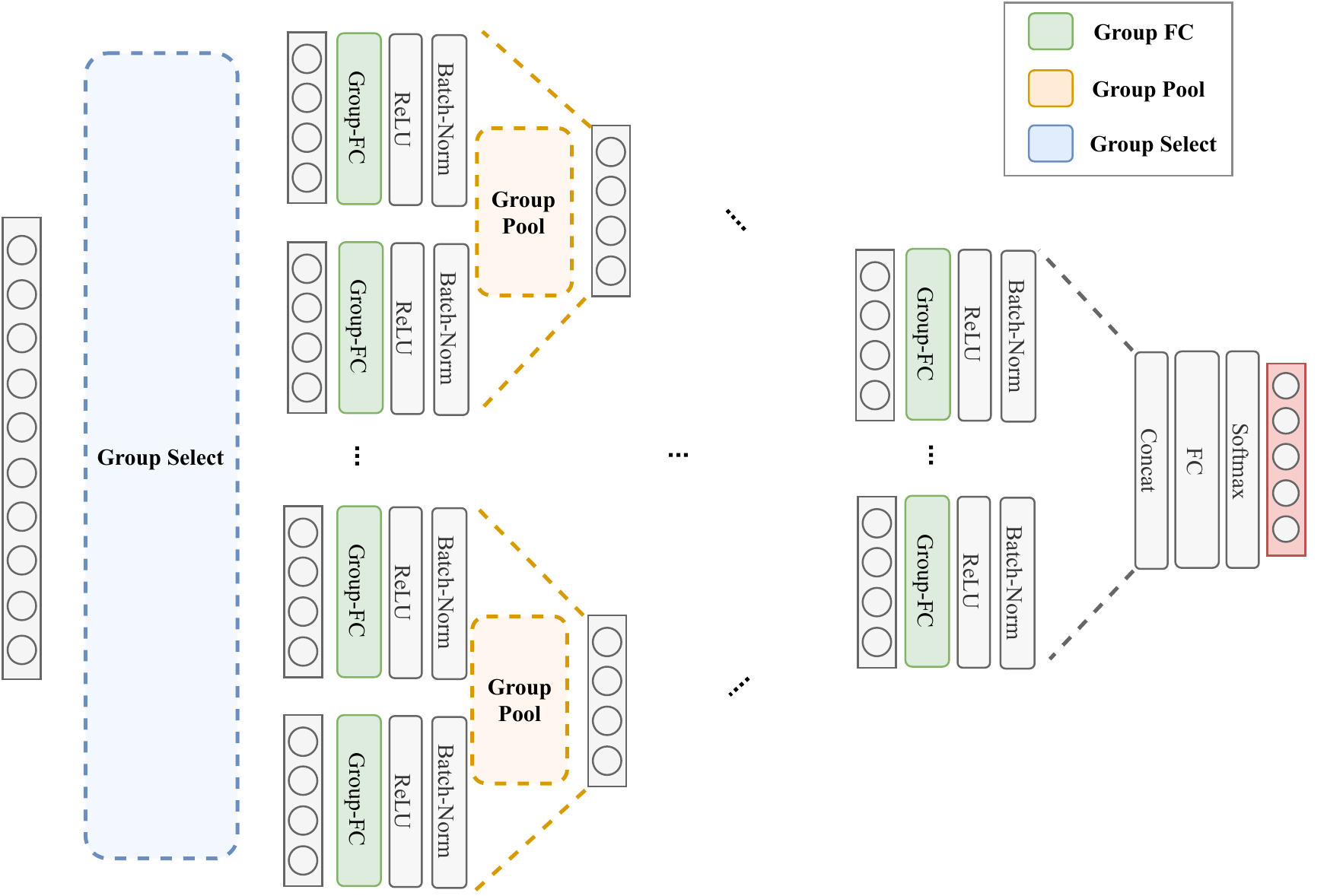}
        \caption{The GMLP network architecture.}
\end{figure*}

\clearpage
\section{Analysis of Intra-Group and Inter-Group Correlations}

\begin{figure}[h]
    \centering
    \begin{subfigure}[b]{0.50\columnwidth}
        \centering
        \includegraphics[width=\columnwidth]{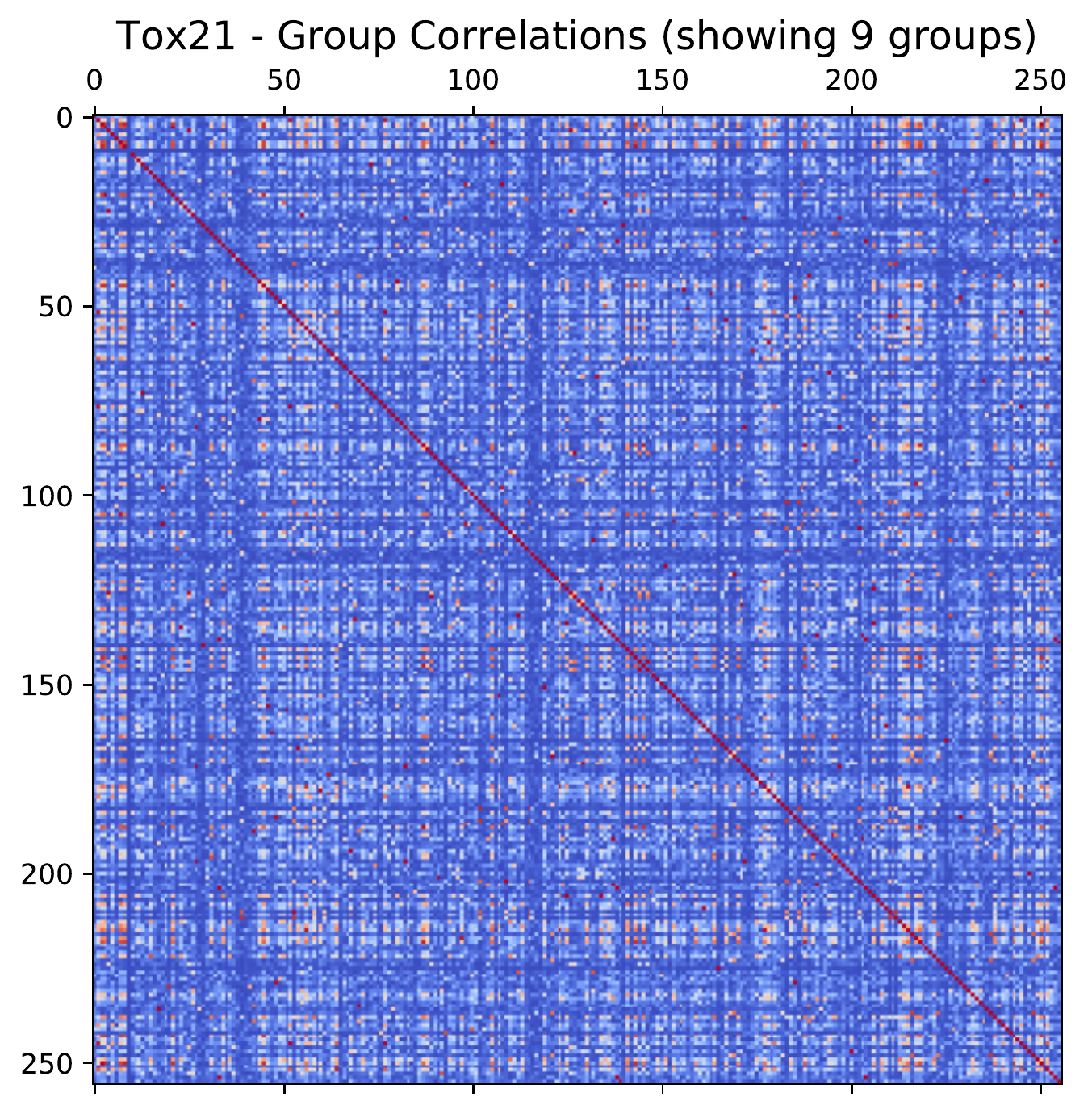}
        \caption{}
    \end{subfigure}
    
    \begin{subfigure}[b]{0.45\columnwidth}
        \centering
        \includegraphics[width=\columnwidth]{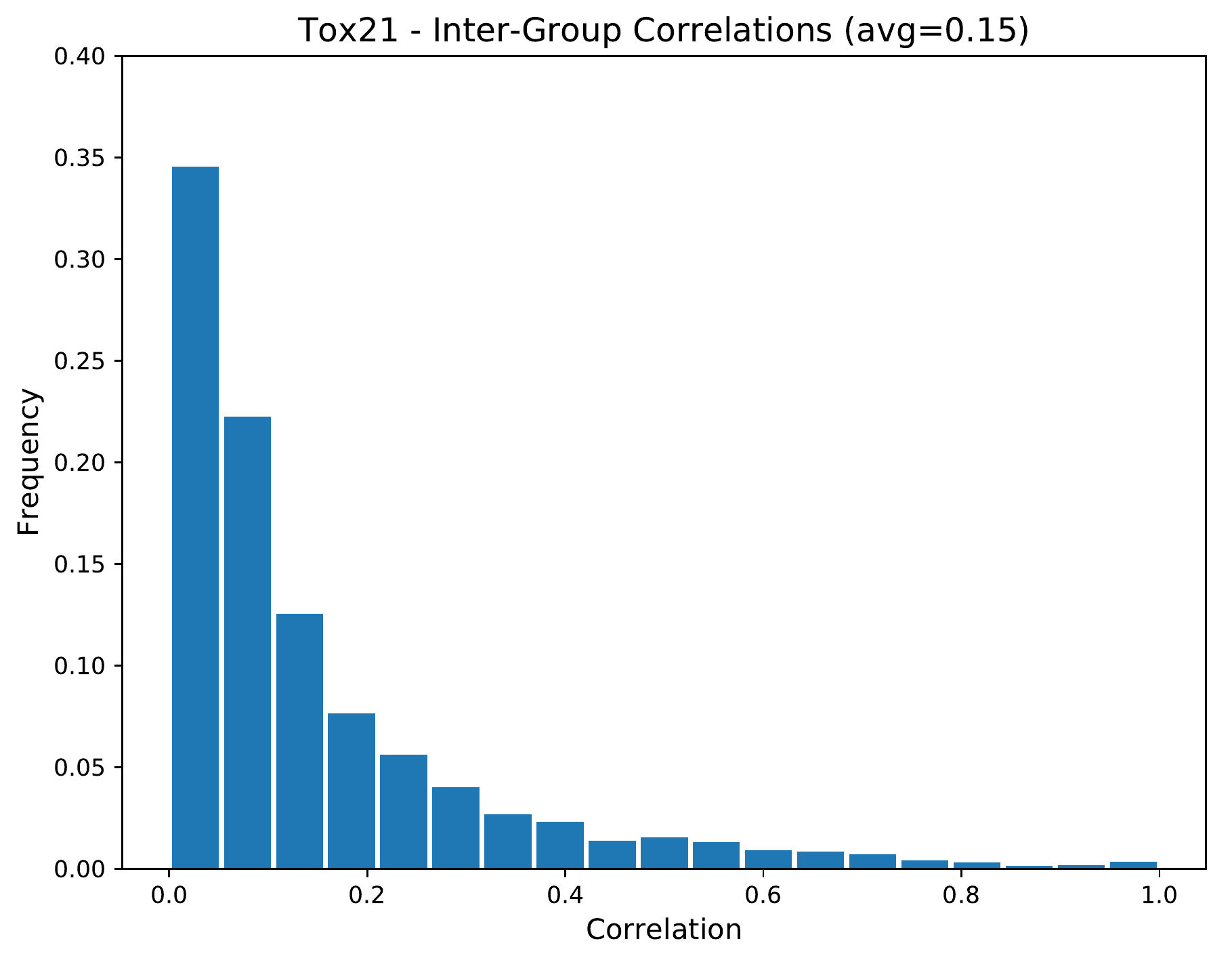}
        \caption{}
    \end{subfigure}
    ~
    \begin{subfigure}[b]{0.45\columnwidth}
        \centering
        \includegraphics[width=\columnwidth]{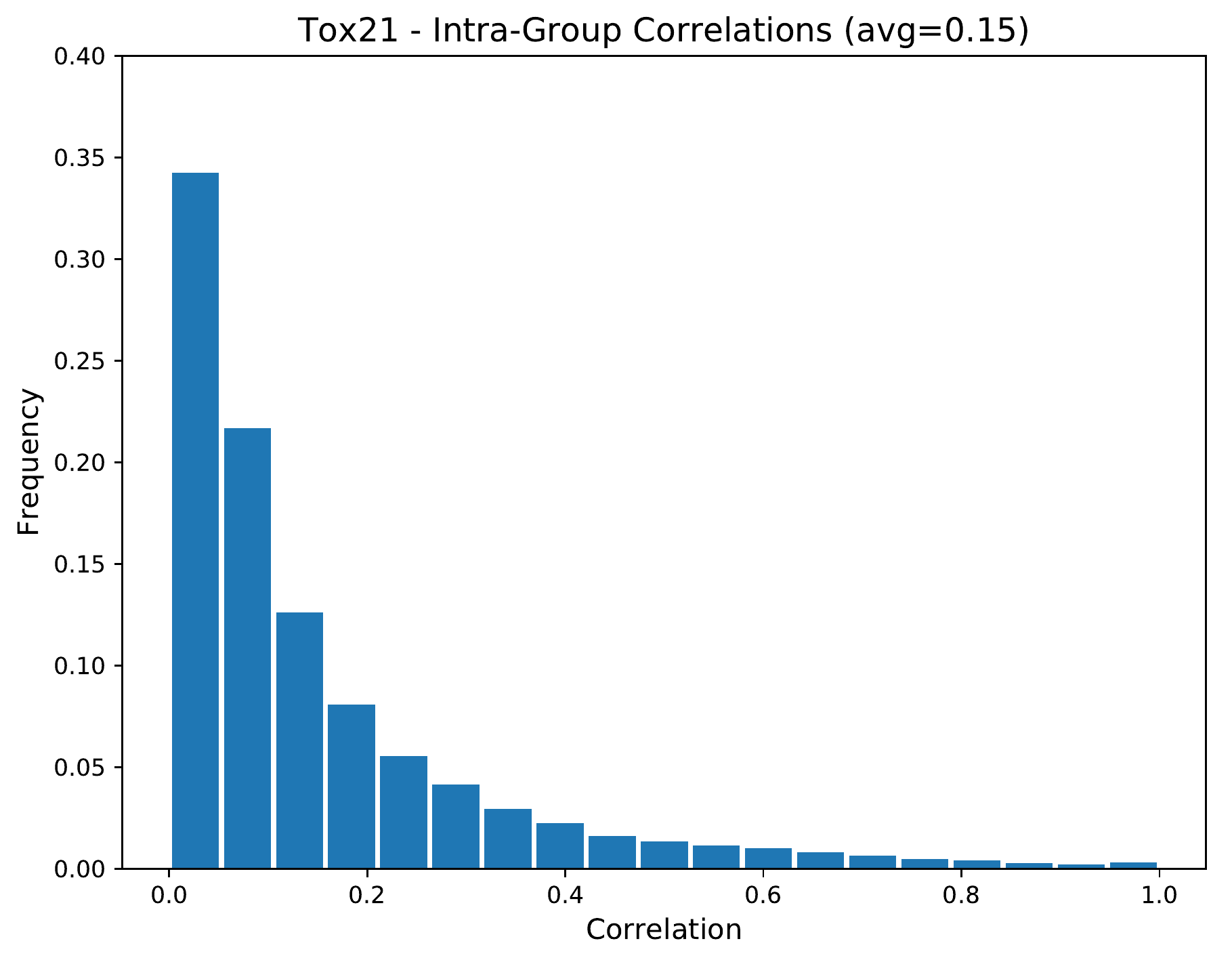}
        \caption{}
    \end{subfigure}
    \caption{Analysis of group correlations for the Tox21 dataset: (a) the correlation matrix for the first 256 output features of the \texttt{Group-Select} layer, (b) the histogram of correlation values computed for inter-group feature pairs, and (c) the histogram of correlation values computed for intra-group feature pairs.}
    \label{fig:corr_tox21}
    \vspace{-0.0in}
\end{figure}

\begin{figure}[h]
    \centering
    \begin{subfigure}[b]{0.50\columnwidth}
        \centering
        \includegraphics[width=\columnwidth]{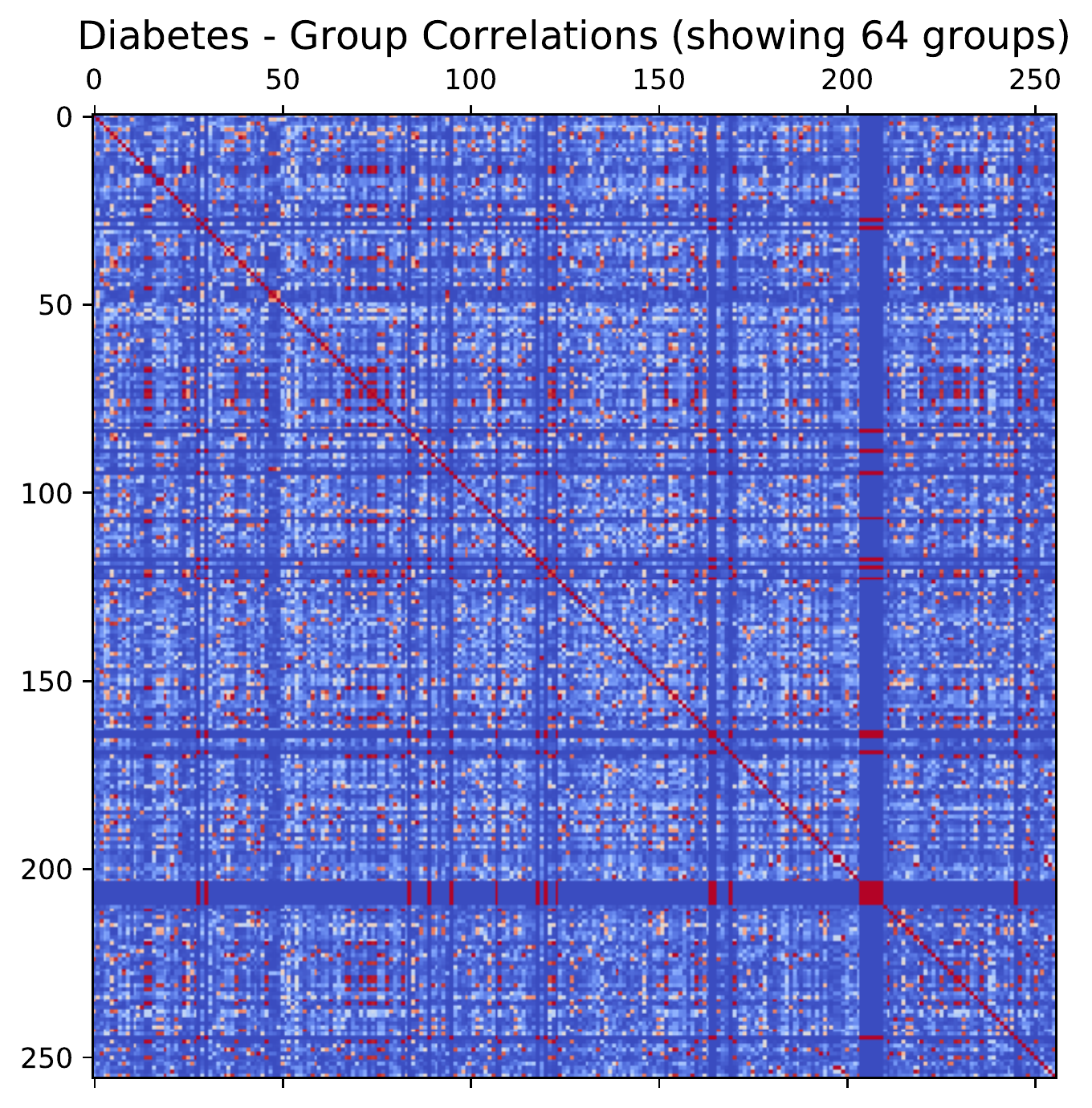}
        \caption{}
    \end{subfigure}
    
    \begin{subfigure}[b]{0.45\columnwidth}
        \centering
        \includegraphics[width=\columnwidth]{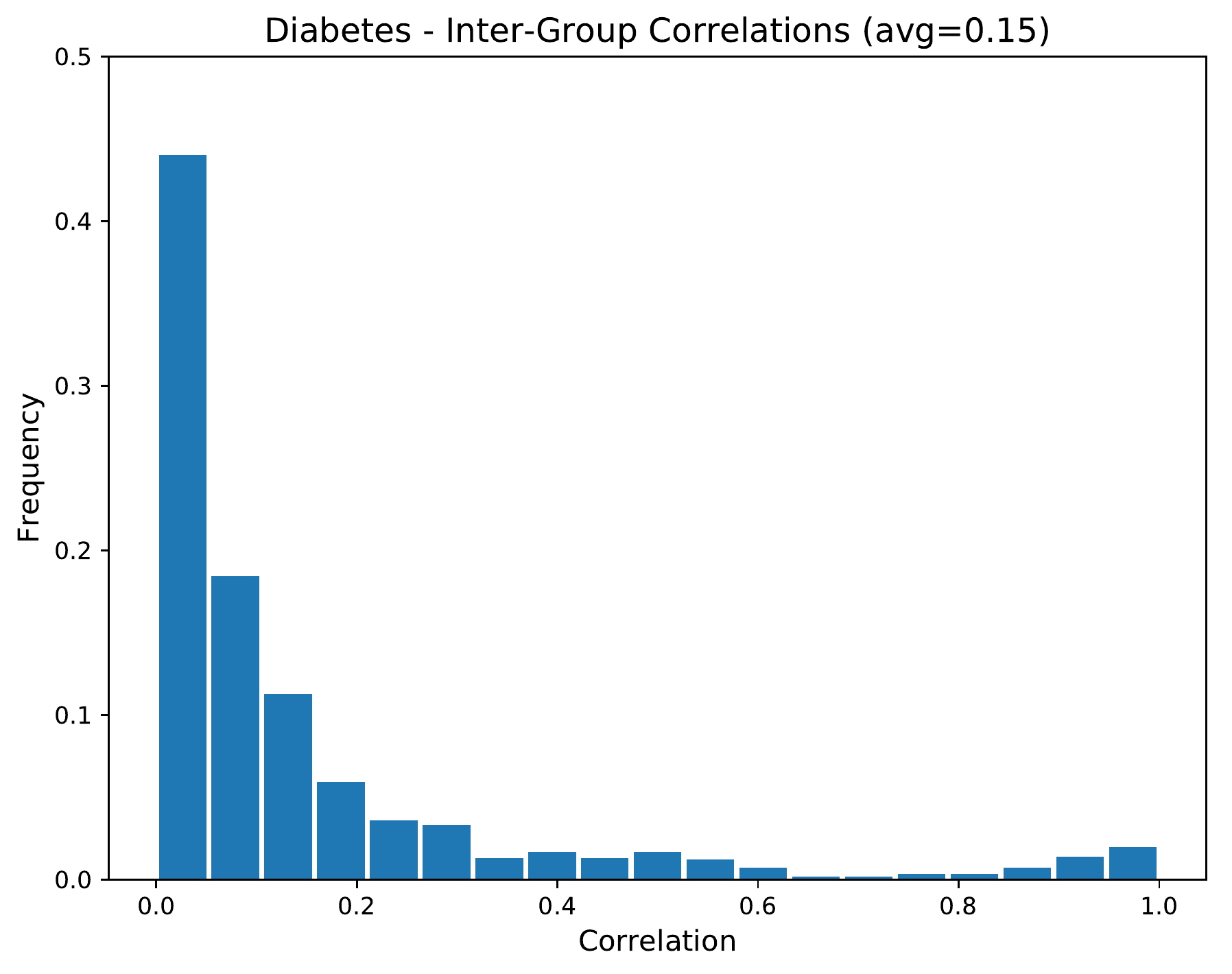}
        \caption{}
    \end{subfigure}
    ~
    \begin{subfigure}[b]{0.45\columnwidth}
        \centering
        \includegraphics[width=\columnwidth]{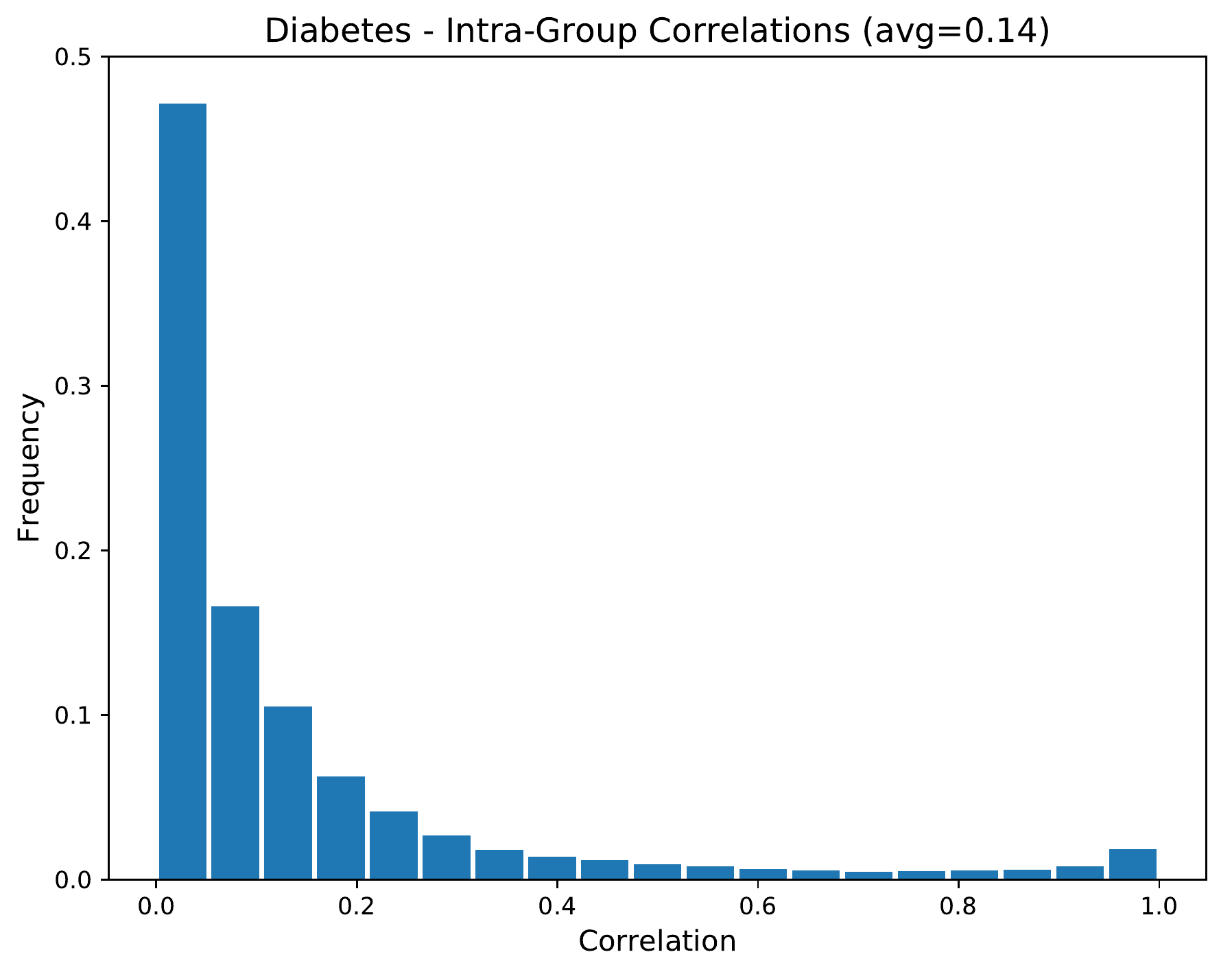}
        \caption{}
    \end{subfigure}
    \caption{Analysis of group correlations for the Diabetes dataset: (a) the correlation matrix for the first 256 output features of the \texttt{Group-Select} layer, (b) the histogram of correlation values computed for inter-group feature pairs, and (c) the histogram of correlation values computed for intra-group feature pairs..}
    \label{fig:corr_diabetes}
    \vspace{-0.0in}
\end{figure}

We conducted experiments using the Tox21 and Diabetes datasets to measure inter-group and intra-group feature correlations. In Figure~\ref{fig:corr_tox21}~and~\ref{fig:corr_diabetes}, subplot (a) shows the correlation matrix for the first 256 output features of the \texttt{Group-Select} layer. Note that the group size is 28 for Tox21 and is 4 for Diabetes. In Figure~\ref{fig:corr_tox21}~and~\ref{fig:corr_diabetes}, subplot (b) and (c) show the histogram of correlation values computed for inter-group and intra-group feature pairs.

From these figures, we do not find any significant difference in the correlation values for features within each group and features between different groups. We believe that this is an expected result as our objective function is based on a classification loss and does not enforce any property among the learned feature groups. Note that often a level of inter-group and intra-group redundancy improves the robustness of the trained models.

    

\clearpage
\section{Comparison of the Softmax and Concrete Relaxations}
In this paper, we used a simple softmax with temperature  to learn the discrete $\Psi$ matrix and implement the \texttt{Group-Select} layer. Alternatively, one can use other ideas such as the concrete distribution suggested by Maddison \textit{et al.}~\footnote{C. J. Maddison, Andriy Mnih, and Yee Whye Teh. The concrete distribution: A continuous relaxation of discrete random variables. In ICLR, 2017.}. For our implementation, we use \texttt{RelaxedOneHotCategorical} class from the PyTorch library that is based on an implementation of the concrete distribution as suggested by Maddison et al. Here, we used a similar temperature annealing schedule and took new samples from the distribution at every forward path computation.

\begin{figure}[h]
    \centering
    \begin{subfigure}[b]{0.45\columnwidth}
        \centering
        \includegraphics[width=\columnwidth]{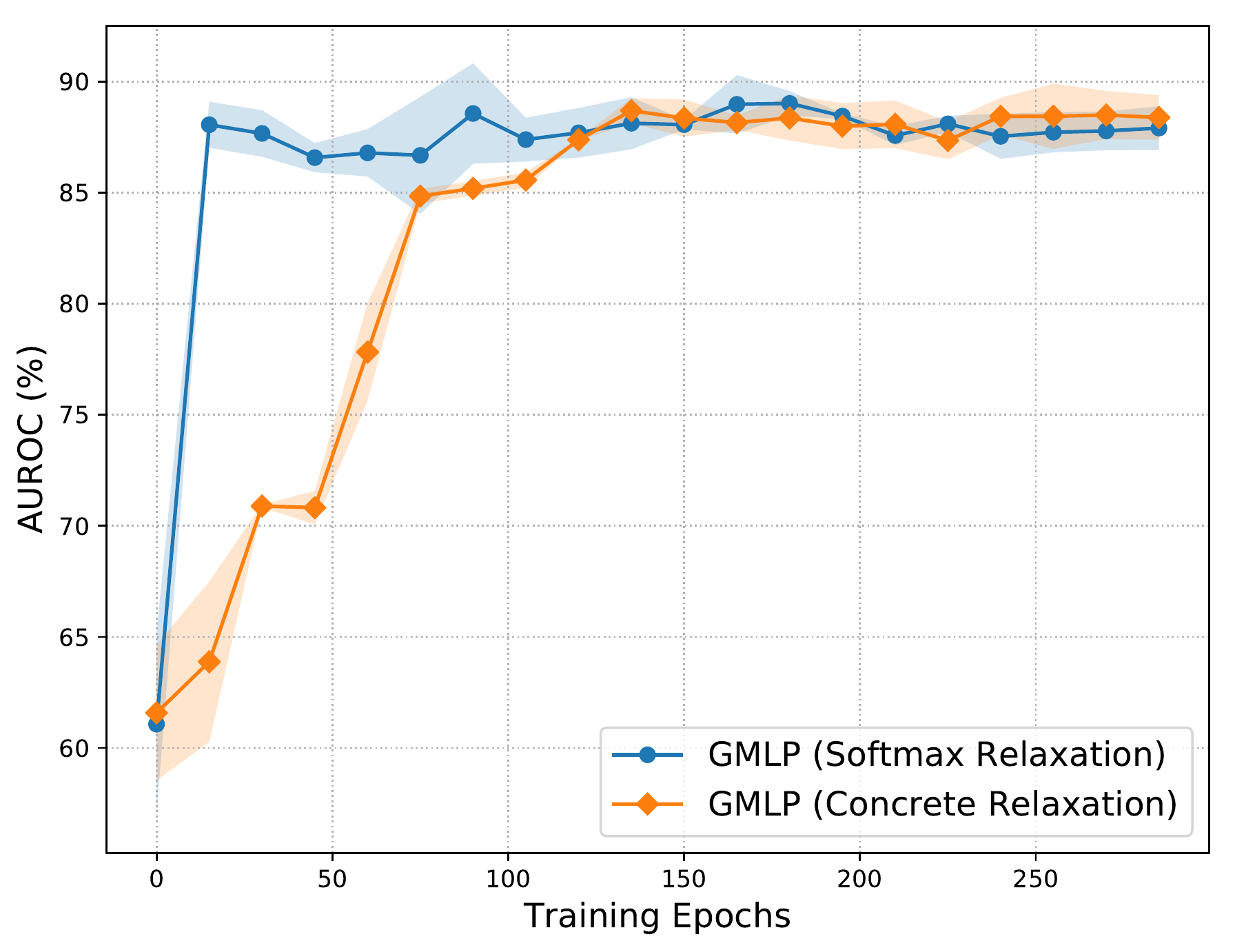}
        \caption{}
    \end{subfigure}
    ~
    \begin{subfigure}[b]{0.45\columnwidth}
        \centering
        \includegraphics[width=\columnwidth]{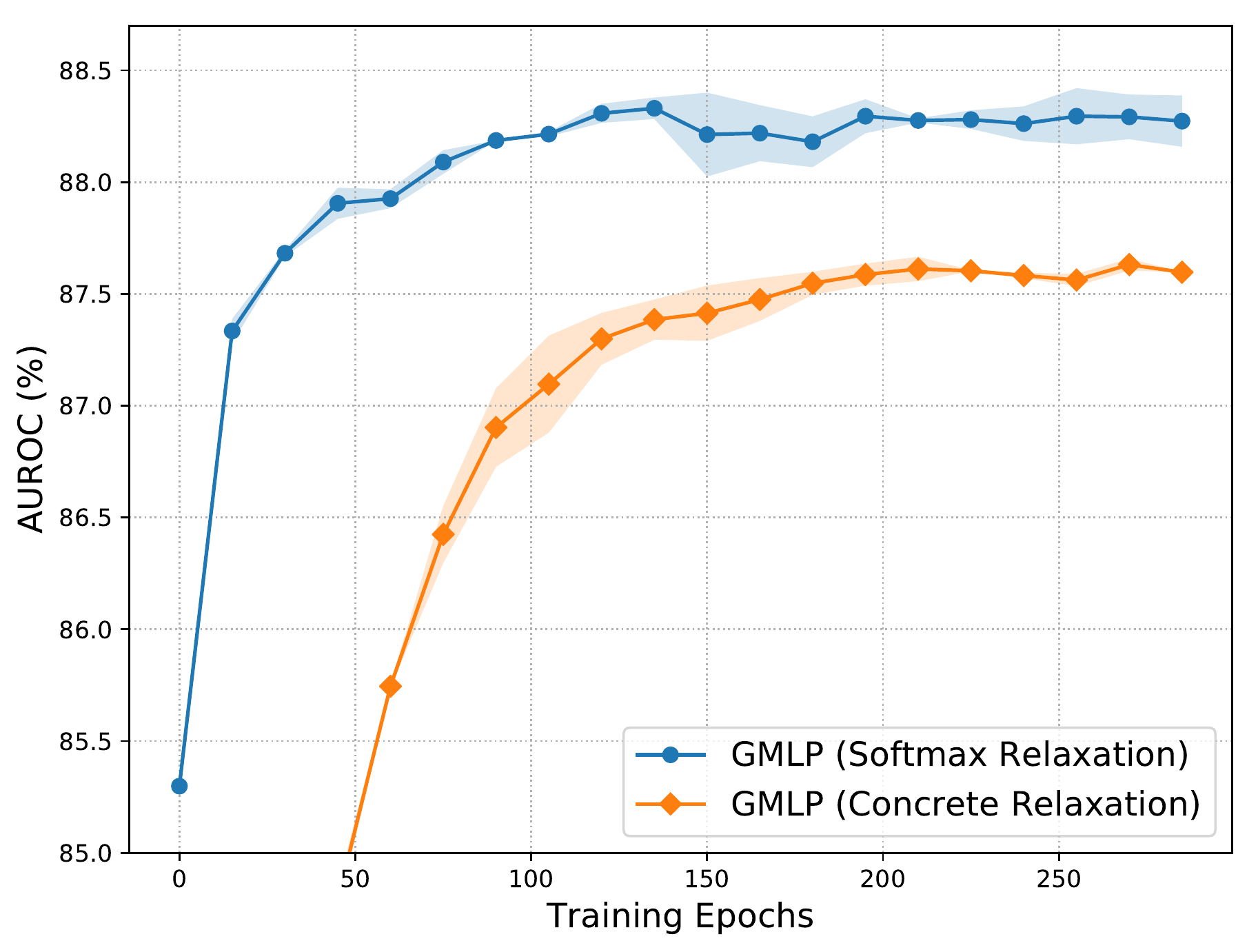}
        \caption{}
    \end{subfigure}
    \caption{Comparison of using the softmax relaxation and concrete relaxation to implement the \texttt{Group-Select} layer: (a) the Tox21 dataset and (b) the Diabetes dataset.}
    \label{fig:ablation_concrete}
    \vspace{-0.0in}
\end{figure}

Figure~\ref{fig:ablation_concrete} provides a comparison of training GMLP networks using the suggested softmax relaxation and the alternative concrete distribution for the Tox21 and Diabetes datasets. Based on this result, we can see that the simpler softmax method achieves similar results for the Tox21 dataset and better results for the Diabetes dataset. We hypothesize that as the major use case of the concrete distribution is in variational methods and it involves random sampling, it might be injecting a level of noise and variation that is not necessarily helpful for learning the feature groups.

\end{document}